\documentclass[journal,twoside,web]{ieeecolor}
\usepackage{generic}
\usepackage{cite}\usepackage{amsmath,amssymb,amsfonts}
\usepackage{algorithmic}
\usepackage{graphicx}
\usepackage{textcomp}
\usepackage{multirow} 
\usepackage{siunitx}
\usepackage{float}
\usepackage{array}
\usepackage{chemformula}
\usepackage{subcaption}
\usepackage{tabularx}
\usepackage[normalem]{ulem}
\usepackage{color}
\usepackage[dvipsnames]{xcolor}
\usepackage{arydshln}
\usepackage{url} 
\def\BibTeX{{\rm B\kern-.05em{\sc i\kern-.025em b}\kern-.08em
    T\kern-.1667em\lower.7ex\hbox{E}\kern-.125emX}}
\newcommand{\revo}[1]{{\color{black} #1}} 
\newcommand{\revt}[1]{{\color{black} #1}} 
\newcommand{\mmrev}[1]{{\color{black} #1}} 
\newcommand{\revf}[1]{{\color{black} #1}} 
\newcommand{\revfi}[1]{{\color{black} #1}} 

\newcommand{\camready}[1]{{\color{black} #1}} 
\newcommand{\final}[1]{{\color{black} #1}} 

\graphicspath{{./figures/}}

\begin{document}


\title{Geometry-Dependent Properties of \\
Physically Crosslinked PPy-PVA Hydrogels \\Enable Tunable Iontophoretic Drug Delivery} 

\author{Heather Jin Hee Kim,~\IEEEmembership{Member,~IEEE}, Mohsen Alishahi, Kyuin Park, Mikaila K. Roncevich, Larissa M. Shepherd, Cara M. Nunez,~\IEEEmembership{Member,~IEEE}
\thanks{Heather Jin Hee Kim and Cara M.\ Nunez are with Sibley School of Mechanical and Aerospace Engineering, Cornell University, Ithaca, NY 14853. (email: jk2768@cornell.edu; cmn97@cornell.edu).}
\thanks{Mohsen Alishahi, Kyuin Park, Mikaila K.\ Roncevich, and Larissa M.\ Shepherd are with Department of Human Centered Design, Cornell University, Ithaca, NY 14853.}
}

\maketitle
\begin{abstract} 
\textit{Objective:} Iontophoresis enables noninvasive transdermal drug delivery. The control of drug release, however, has largely relied on current modulation. Here, we explore geometry-driven factors, gel microstructure, and local electric fields, as alternative levers for regulating drug transport. 
\textit{Methods:} \revo{To investigate iontophoretic drug delivery, we encapsulate bovine serum albumin (BSA) in polypyrrole-polyvinyl alcohol (PPy-PVA) hydrogels. Five distinct hydrogel geometries were fabricated to impose variation in pore architecture and electric field distribution, including Hilbert curve-based designs that use increased corner density to enhance local electric field strength. Using formulation studies, we determined the ideal PPy content, drug loading efficiency, and current density. Maximum field strength of individual hydrogel geometries was simulated via finite element analysis, and microstructures were characterized by scanning electron microscopy.} 
\revo{Lastly, we assessed differences in the transport metrics of the hydrogel geometries through an in vitro study using a Franz diffusion cell} \mmrev{and an artificial membrane}. 
\textit{Results:} The engineered geometries produced clearly differentiated microstructures and electric field strengths. Drug delivery studies demonstrated that iontophoresis substantially enhanced permeability compared with passive diffusion, with each geometry yielding \revfi{different} profiles in cumulative release, flux, and permeability. Notably, the average iontophoretic permeability was \camready{$(3.80\,\pm \,1.93)\times10^{-5}$}, exceeding values reported \revf{in previous studies using artificial membranes.} \textit{Conclusion:} This study showed that the geometry-driven fabrication of drug-loaded hydrogel can achieve \revfi{different} delivery characteristics in both iontophoretic and passive transport \revf{using artificial membranes}. \textit{Significance:} Geometry-mediated control complements existing approaches for iontophoretic drug delivery without requiring individual current-density control. 

\end{abstract}
\section{Introduction}

\subsection{Motivation}

\begin{figure}[b]
	\centering
	\includegraphics[width=\linewidth]{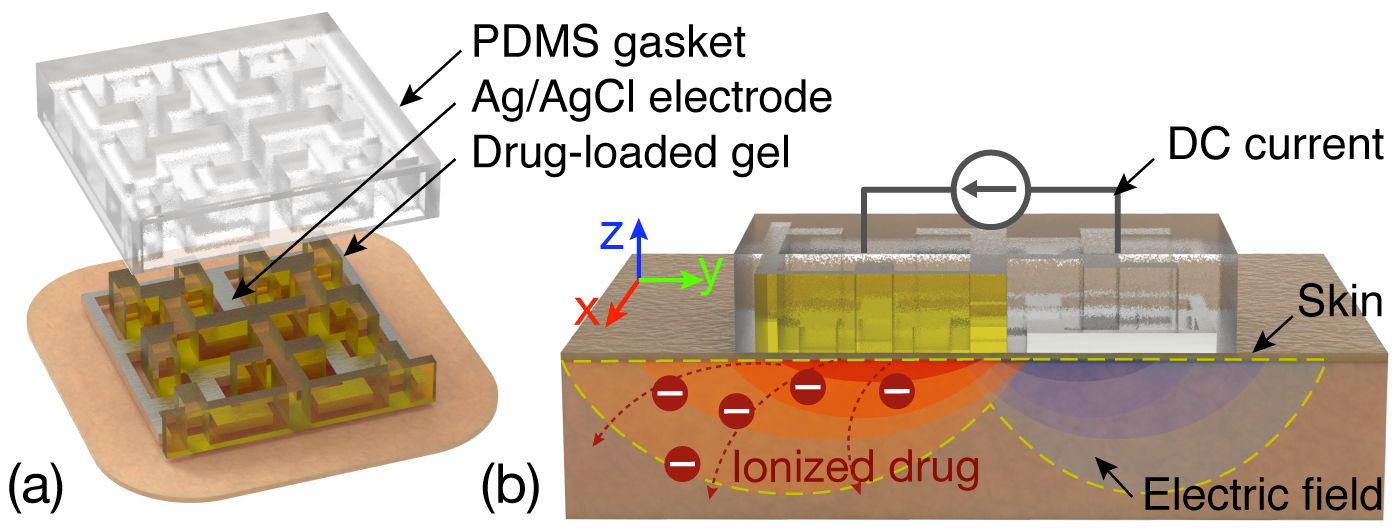}
	\caption{(a) Exploded view of the device. The device integrates a drug-loaded hydrogel, Ag/AgCl electrodes, and a PDMS gasket. (b) When current is applied, the resulting electric field drives the transport of ionized drug molecules through the skin barrier.} \label{fig:teaser} 
\end{figure}

Transdermal drug delivery, the delivery of ionized drug molecules through the layers of the skin, has gained increased attention due to its minimally invasive administration, potential for reduced skin side effects, and improved patient adherence to treatment~\cite{shah_InnovativeTransdermal_2025}. The primary challenge of transdermal drug delivery is penetrating the stratum corneum, the outermost layer of the epidermis. This layer is composed of keratinized corneocytes embedded in a dense lipid matrix and serves as a formidable barrier to most therapeutic molecules~\cite{bouwstra_SkinBarrier_2023, prausnitz_SkinBarrier_}. For large drug molecules or other peptide-based therapeutics, the stratum corneum significantly hinders passive delivery.

Our work focuses on iontophoresis, which applies a mild constant or pulsed current to drive ionized drug molecules across the skin via an imposed electric field~\cite{costelloIontophoresisApplicationsTransdermal1995} (Fig. \ref{fig:teaser}). The required current density varies depending on the delivery medium and the study context (e.g., in vitro or in vivo), and has been reported in the literature to range from as low as 5 $\mu\mathrm{A\,cm^{-2}}$ to as high as 3 mA\,cm$^{-2}$~\cite{yangSmartphonepoweredIontophoresismicroneedleArray2020}. A typical iontophoretic drug delivery system consists of a power source, a pair of electrodes serving as the anode and cathode, and a drug reservoir, which is typically a gel or solution containing the active drug placed against the skin. 
Upon generation of an electric field from the applied current, charged drug molecules are repelled from the electrode of like charge and driven through the stratum corneum. The kinetics of iontophoretic delivery follow the Nernst-Planck equation~\cite{kastingTheoreticalModelsIontophoretic1992}, where the flux is governed by both electrochemical gradients and electric field strength. \revo{Particularly, chemically or physically crosslinked hydrogel-based iontophoresis has been successfully employed to deliver a wide range of therapeutics}~\cite{helmyOverviewRecentAdvancements2021}, including bovine serum albumin (BSA)~\cite{zhao_HydrogelIonic_2022, wei_ApplicationHydrogels_2023}, lidocaine~\cite{wangLidocaineLoadedIontophoresisDrivenFiberBased2025}, insulin~\cite{zakzewskiTransdermalDeliveryRegular1998, pillaiTransdermalDeliveryInsulin2003, kanikkannanTransdermalIontophoreticDelivery1999, pengIontophoresisIntegratedSmartMicroneedle2024, moraradIontophoresisBasalInsulin2022, yangSmartphonepoweredIontophoresismicroneedleArray2020}, and vaccines~\cite{zhengIontophoresisdrivenMicroneedlePatch2023}. 
Iontophoretic delivery using gels enables fine control over release kinetics by tuning parameters such as current density, waveform, and pulsing schemes~\cite{bhatiaEffectModulatedAlternating2014}.  
While this approach offers substantial flexibility, most strategies for controlled release have focused on modulating electrical input, whereas alternative routes, such as tailoring the gel’s permeability or structural properties, remain comparatively underexplored. \revo{To further expand the controllability of drug release, this work is motivated by the freeze-thaw process of physically crosslinked hydrogels, where gel properties are highly sensitive to geometry due to mold-induced mechanical stress. We hypothesize that mold-induced hydrogel geometry can act as an indirect mediator that modulates key transport-relevant properties of drug-loaded hydrogels, namely their porous microstructure and local electric field strength, both of which appear in the governing equations for iontophoretic transport.} 

\revo{In this work, we conduct an in vitro study to investigate how hydrogel geometry serves as an indirect modulator of drug release behavior by influencing both microstructure and electric field strength during freeze-thaw-induced physical crosslinking. Using bovine serum albumin (BSA) as a model drug, we investigate how geometry-driven variations in the gel affect (1) microstructure and (2) electric field distribution, and how these factors collectively govern drug release. To this end, we design five distinct hydrogel geometries using mold-defined templates with varied shapes (but consistent area) to generate a range of electric field intensities. Because freeze-thaw processing inherently couples geometry with microstructural evolution, each geometry produces a unique microstructure arising from mechanical stresses experienced during crosslinking within the mold  (Section~\ref{sec:background_5}).} The gel was formulated using a one-factor-at-a-time process across all geometries (Section~\ref{sec:cd_result}). We then conducted a drug delivery study to assess whether transport parameters differ across gel geometries (Section~\ref{sec:dd_method}). In Section~\ref{sec:dd_result}, we present results, which show that geometry-mediated variations in pore structure and electric-field localization influence iontophoretic transport outcomes.

\vspace{-.8em}

\subsection{Background}

\subsubsection{Governing Kinetics of Iontophoretic Transport}\label{sec:pnpmodel}

The governing equations for iontophoresis are the Poisson-Nernst-Planck (PNP) equations, which couple ionic transport and electrostatics. 
The Nernst-Planck equation describes the flux, $J_i$, of species $i$ as the combined result of (1) diffusion due to a concentration gradient ($-D_i \nabla c_i$), (2) electromigration under an electric potential gradient ($\frac{z_i D_i F}{RT} \, c_i \nabla \phi$), and (3) convection driven by the fluid velocity ($c_i\mathbf{v}$) (Eq.\ref{eq:NP}). In this expression, $D_i$ is the diffusion coefficient of species $i$, $\nabla c_i$ the gradient of the concentration of species $i$, $z_i$ is the valence of species $i$, $F$ is Faraday's constant, $R$ is the universal gas constant, $T$ is the absolute temperature, $\nabla \phi$ is the electric potential gradient, and $\mathbf{v}$ is bulk fluid velocity vector.  

\begin{equation} \label{eq:NP}
     J_i = -D_i \nabla c_i \;-\; \frac{z_i D_i F}{RT} \, c_i \nabla \phi \;+\; c_i \mathbf{v} 
\end{equation}

Complementing this, the Poisson equation (Eq.~\ref{eq:poisson}) expresses Gauss’s law in differential form (Eq.~\ref{eq:gauss_diff}) by relating the divergence of the electric displacement field to the local charge density, where $\phi$ is the electric potential, $\varepsilon$ is the permittivity of the medium, and $\rho$ is the charge density. Together, the equations couple ionic motion with the resulting electric field. Poisson’s law determines the electric field from charge distribution, while the Nernst-Planck equation describes how ions respond to that field.
\begin{equation} \label{eq:poisson}
    -\nabla \cdot \left( \varepsilon \nabla \phi \right) = \rho = F \sum_i z_i c_i
\end{equation}
\begin{equation} \label{eq:gauss_diff}
    \mathbf{E} = -\nabla \phi
\end{equation}

\noindent For a planar surface, Gauss’s law (Eq.~\ref{eq:gauss}) gives the relation between the electric field, $E$, and surface charge density, $\sigma$, as:
\begin{equation} \label{eq:gauss}
    \mathbf{E} = \frac{\sigma}{\varepsilon_{n}}
\end{equation}

\noindent where $\varepsilon_{n}$ is the normal component of the permittivity at the surface. 


\subsubsection{Local Enhancement of Electric Fields From Electrode Geometry}

Optimization of an electrode's electric field strength is critical across a range of applications, including electroadhesion~\cite{guo_optimization_2016} and field-responsive gas sensor systems~\cite{yang_optimizing_2021}. In these prior works, electrodes were deliberately engineered with regions of high charge density, such as sharp corners, edges, or areas of pronounced curvature, to generate locally intensified electric fields, thereby enhancing electroadhesive forces and improving sensor sensitivity. Following Gauss’s law (Eq.~\ref{eq:gauss}), electric-field lines converge and emerge perpendicularly from these geometrical features, resulting in pronounced field amplification near tips and edges. Such localized field enhancement is expected to play a similarly important role in influencing iontophoretic drug transport.  

To exploit this electrostatic phenomenon, we employ parametrically generated Hilbert-Piano curves in our electrode geometries. Hilbert-Piano curves are a class of space-filling fractal curves characterized by recursive geometric complexity (Supplementary Fig.~S1). 
Hilbert-Piano curves generate a large number of corners and sharp angles, which can be harnessed to concentrate electric fields and amplify local field intensity in electrode designs~\cite{yang_optimizing_2021, guo_optimization_2016}.

\subsubsection{Polypyrrole-Polyvinyl Alcohol Hydrogel and Bovine Serum Albumin}

\revo{Hydrogels play a central role in iontophoretic drug delivery systems, however, the hydrogel itself must be ionically conductive. Therefore, among the available materials that enable this, we use conductive polypyrrole (PPy) as it stands out not only for its intrinsic electronic conductivity, but also its chemical stability, environmental robustness, and biocompatibility~\cite{wu_BiomedicalApplications_2024, han_FlexibleConductive_2022, parida_HighlyPorous_2023}. When doped, PPy supports mobile charge carriers along its conjugated backbone, making it especially attractive for electroactive drug-delivery platforms. Blending PPy with polyvinyl alcohol (PVA) yields a composite hydrogel that is both mechanically resilient and ionically conductive, an ideal environment for loading macromolecules such as bovine serum albumin (BSA).}

\revo{BSA is a globular protein consisting of 583 amino acids with a molecular weight of approximately 66.5 kDa~\cite{meng_PreparationDrugLoaded_2022}. BSA is frequently employed as a model macromolecular drug in transdermal and nanoparticle-based delivery studies, where it often serves as a surrogate for other large therapeutic proteins, such as insulin~\cite{ganguly_BovineSerum_2024, meng_ResearchProgress_2023}. Because BSA shares structural and functional similarities with human serum albumin, it enables its use in pharmacokinetic modeling and drug-protein interaction research without the cost or ethical concerns associated with human-derived materials~\cite{aguilera-garrido_ApplicationsSerum_2021}. Furthermore, BSA-based nanocarriers are among the most widely investigated drug delivery platforms due to their stability, solubility, and efficient uptake into tissues~\cite{karimi_AlbuminNanostructures_2016}}. 

\subsubsection{Impact of Hydrogel Preparation on Microstructure} \label{sec:background_5} 

The freeze-thaw (F/T) technique is one of the most widely used physical crosslinking methods and is particularly attractive because it generates porous networks without chemical crosslinkers~\cite{liang_PolyvinylAlcohol_2024}. In this process, a precursor polymer solution is subjected to repeated cycles of F/T. 
These physically crosslinked regions persist upon thawing, producing a stable, semi-crystalline network~\cite{holloway_RoleCrystallization_2013}. \revo{During freezing, ice formation induces volumetric expansion, so gels confined in rigid molds develop significant internal stresses as the phase change progresses. In F/T hydrogels, ice crystallization and melting act as a porogen, with fluid-matrix interactions and volume changes reshaping pore size, connectivity, and network morphology. Under rigid confinement and consistent F/T conditions, these stresses and local expansions strongly govern pore nucleation, alignment, and evolution within the gel network~\cite{levin_SwellingConstraints_2023}. The role of mold-induced geometric confinement in shaping microstructure, permeability, and drug delivery performance remains underexplored relative to other F/T parameters~\cite{gupta_CombinedEffect_2012}.}

\vspace{-.5em}

\section{Materials and Methods}



\subsection{Calibration and Spectral Component Separation}

\subsubsection{Spectral Unmixing of PPy and BSA} \label{sec:nnls}

Most PPy nanoparticles in the gel composite are immobilized within the PVA hydrogel matrix through physical entanglement and hydrogen bonding. Nevertheless, a small fraction of unbound nanoparticles may undergo passive diffusion through the gel network. 
In such cases, spectral overlap can occur between PPy and BSA, as PPy exhibits a broad absorption peak spanning 250-300 nm~\cite{elkomy_StructuralOptical_2016, pagar_SynthesisPhysical_2023, zhao_NonlinearOptical_2017}, whereas BSA displays a pronounced absorption peak near 278 nm~\cite{bronze-uhle_synthetic_2016}. To enable separation of their signals, we apply spectral unmixing using non-negative least squares (NNLS) regression~\cite{keshava_survey_2003}. NNLS separates overlapping signals by estimating component ratios (amplitudes) that cannot be negative. NNLS is a widely used and well-established regression technique for unmixing overlapping spectra across modalities, including ultraviolet-visible spectroscopy (UV-Vis)~\cite{zobel-roos_process_2017, steiert_photophysical_2018}, fluorescence~\cite{steiert_photophysical_2018}, and Raman spectroscopy~\cite{zhang_MixtureAnalysis_2014}, as well as fluorescent microscopic imaging~\cite{orth_super-multiplexed_nodate, pengo_efficient_2013}. Using NNLS, the observed spectrum of our sample from the receptor medium is expressed as a non-negative linear combination of the reference spectra of PPy and BSA, plus a polynomial baseline. After setting target wavelength window $[\,\lambda_{\text{min}}, \lambda_{\text{max}}\,]$, for each observed spectrum $A(\lambda)$, we model:
\begin{equation}
A(\lambda) \;\approx\; a_{B} S_{B}(\lambda) \;+\; a_{P} S_{P}(\lambda) \;+\; \beta_0 + \beta_1 \lambda + \beta_2 \lambda^2
\end{equation}

\noindent where $a_B,a_P \geq 0$ are the BSA and PPy contributions, and $\beta_0,\beta_1, \text{and} \,\beta_2$ are unconstrained baseline terms. $S_B$ and $S_P$ are averaged reference spectra from BSA-only and PPy-only surrogate standard samples. To prevent the broad PPy band from confounding baseline fitting, the polynomial baseline matrix $P=[1,\lambda,\lambda^2]$ was orthogonalized relative to the spectral subspace $Q=[S_B,S_P] \in \mathbb{R}^{n \times 2}$, resulting in $P_{\perp}$ in Eq.~\ref{eq:ortho}. Orthogonalization turns spectral basis vectors of BSA, $S_B$, and PPy, $S_P$, into an orthogonal set such that each basis spectrum becomes perpendicular, and thereby linearly independent.
\begin{equation} \label{eq:ortho}
P_{\perp} \;=\; P - Q(Q^\top Q)^{-1} Q^\top P
\end{equation}

\noindent The resulting final regression problem is:
\begin{equation}
    \min_{c\in\mathbb{R}^5}\ \|Xc - A\|_2^2\quad\text{s.t.}\quad c_1 \ge 0,\ c_2 \ge 0   
\end{equation}
with $X=[S_B, S_P, P_{\perp}]$ and $c=[a_B,a_P,\beta_0,\beta_1,\beta_2]^\top$, where $c_1$, and $c_2$ denote non-negative amounts of BSA, $a_B$, and PPy, $a_P$, present in the mixture. These coefficients represent the strengths of the reference spectra, $S_B, \text{and} \, S_P$, required to reconstruct the measured sample spectrum. We used the convex NNLS solver with two fallback routines (MATLAB 2025b). The obtained amplitudes $a_B \, \text{and} \, a_P$ were converted to concentrations using the respective calibration curves described in the following paragraphs. NNLS was applied in Sections~\ref{sec:ee} and~\ref{sec:cd} to unmix the unknown concentrations of BSA and PPy from the measured spectra.



\paragraph{BSA Calibration Curve}\label{sec:bsa_cc} We first constructed a single wavelength calibration curve using a UV-Vis spectrometer (Cary 5000). BSA was dissolved in PBS at various concentrations and prepared in quartz cuvettes. For BSA, we observed $\lambda_{max}$ at 278 nm, which is in agreement with the literature~\cite{arya_quantitative_nodate, bronze-uhle_synthetic_2016}. To map this NNLS-derived BSA amplitude, $a_B$, to concentration, the BSA reference spectrum, $S_B$, was normalized such that $S_B(278) = 1$. This step guarantees that the fitted amplitude, $a_B$, represents the absorbance of BSA at 278 nm and can be directly related to the Beer-Lambert calibration curve. Because BSA exhibits a strong and specific absorption peak, a single-wavelength calibration was appropriate. The obtained normalized amplitudes, $a_B(278)$, were then plotted against the known concentrations of BSA (Supplementary Fig.~\final{S2}(a)). The resulting relationship was linear, with R$^2$ value 0.9912, and the curve was $a_B(278) = (5.25 \times 10^{-4}) \, C_{B} + 1.11\times 10^{-3}$, where $C_B$ is BSA concentration in $\mu$g/mL. 



\paragraph{PPy Calibration Curve} Unlike BSA, whose narrow absorption peak near 278 nm enables single wavelength Beer-Lambert calibration, PPy exhibits a broad overlapping UV spectrum, requiring the entire spectrum to be used for calibration. To quantify PPy, we prepared PPy-PBS surrogate standards at known concentrations, $C_s$, for each standard sample, $s$, and estimated the PPy amplitude, $a_P$, by solving a one-analyte NNLS with a free quadratic baseline over the same spectral window used for delivery fits (PPy basis, $S_P$, and baseline columns, $P_\perp$, orthogonalized to $S_P$): 
\begin{equation}
    \min_{a_P^{(s)}\ge 0,\ \beta^{(s)}\in \mathbb{R}^3}\ \left\|\,a_P^{(s)}\, S_P \;+\; P_{\perp}\beta^{(s)} \;-\; A^{(s)}\,\right\|_2^2
\end{equation}

The resulting amplitudes, $a_P^{(s)}$, were mapped to the known concentrations, $C_s$, by ordinary least squares values: 
\begin{equation}
    C_s \;=\; \alpha\,a_P^{(s)} \;+\; \beta
\end{equation}

\noindent Where the slope, $\alpha$, converts the amplitude scale into concentration units, and the intercept, $\beta$, shifts the calibration curve to match the data.
Goodness-of-fit is quantified by:
\begin{equation}
    R^2 \;=\; 1 - \dfrac{\sum_s\big(C_s-(\alpha \, a_P^{(s)}+\beta)\big)^2}{\sum_s\big(C_s-\overline{C}\big)^2}
\end{equation}

The resulting calibration, $C_P = 0.7667 * a_P-0.4886$, achieved agreement R$^2$ = 0.9942 (Supplementary Fig. S2(b)), where $C_P$ is PPy concentration in vol\%. The same mapping was later used to convert NNLS-derived PPy amplitudes from delivery spectra into concentration values (Section~\ref{sec:dd_result}). The range of calibration points, $\{ a_P^{(s)}, \, C_s \}$, established the empirical linearity limits used for pre-dilution screening. 

\paragraph{Experimental Validation of NNLS Model} 
Before the main delivery study, we performed a preliminary iontophoretic experiment using a gel prepared with the same formulation described in Section~\ref{sec:devicedesign}. The gel was subjected to a constant current of 2 mA for 8 hours. 
Samples were collected at 0, 15, 30, 60, 120, 240, and 480 minutes. The resulting spectra were analyzed using the NNLS model. To experimentally validate the model’s fit, we reconstructed spectra at 15, 120, and 240 minutes by preparing samples according to the NNLS-estimated concentrations of PPy and BSA. The reconstructed spectra were compared with the absorbance values predicted by the NNLS algorithm. Across these validation points, the standard deviation of the differences between measured and predicted absorbance at 278 nm was $0.044$~AU, indicating modest scatter of the validation data around the NNLS fit. 


\subsection{Iontophoresis Device Design} \label{sec:devicedesign}

\subsubsection{Materials}

\revo{Pyrrole, polyvinyl alcohol (PVA, Mw 89,000-98,000), hydrogen peroxide (H$_2$O$_2$, 30~\%), deionized (DI) water, sodium dodecyl sulfate (SDS), phosphate-buffered saline (PBS, 1x, pH 7.4), and bovine serum albumin (BSA) were purchased and used as received from Sigma Aldrich. DI water was used for all solution preparations. Whatman filter paper (0.45 $\mu$m pore size) was obtained from Lab Source. Silver/silver chloride (Ag/AgCl) paste for electrodes was sourced from Voltera. Polydimethylsiloxane (PDMS) was obtained from Dow Corning.}

\subsubsection{Preparation of Drug-Loaded PPy-PVA Hydrogel}


The polypyrrole nanoparticle (PPy-NP) solution was prepared by adding 2.9 mmol of pyrrole and 2.9 mmol of SDS to 35 mL of DI water~\cite{tari_controlled_2021}. 
The mixture was ultrasonicated for 30 minutes to ensure proper dispersion. Subsequently, 30.9 mmol of hydrogen peroxide 
was gradually introduced as the oxidizing agent. The initial solution exhibited a faint yellow tint due to the presence of pyrrole. The reaction mixture was then stirred at room temperature using a magnetic stirrer. After 20 hours of stirring, the solution turned dark green, indicating the formation of polymerized PPy, although the resulting PPy at this stage was insoluble (Supplementary Fig.~{S3}(a)).

To obtain water-soluble PPy nanoparticles (PPy-NPs), the reaction mixture was stirred at 80 $^\circ$C for 3 hours, then cooled to room temperature. Almost immediately upon heating, the solution changed to a brown color, indicating the progression of the reaction. The final solution was filtered using a Whatman filter 
to remove particles exceeding the pore size, yielding a stable dispersion of PPy-NPs. Simultaneously, 13 wt\% of polyvinyl alcohol 
was dissolved in DI water for 24 hours at 80 $^\circ$C until the solution achieved clear transparency, and cooled down to room temperature (Supplementary Fig.~\final{S3}(b)). The PPy-NP solution is then blended with the PVA solution under continuous stirring for 5 hours to ensure homogenous mixing. Upon blending, the resulting PPy-PVA mixture exhibits increased viscosity. 
To load BSA into the prepared PVA-PPy solution, BSA was first dissolved in PBS 
at a concentration of 20 mg\,mL$^{-1}$  (Supplementary Fig.~\final{S3}(c)). This concentration was selected based on the entrapment efficiency results described in Section~\ref{sec:ee}, which identified the most effective drug loading conditions. The resulting BSA solution was then added to the PVA-PPy solution and stirred for 1 hour. The resulting solution is then cast into molds (Supplementary Fig.~\final{S3}(d)) corresponding to 5 electrode geometries (Section~\ref{sec:geometry}). \revo{The solution fills the mold’s negative space, generating three-dimensional structures with uniform height, resulting in consistent volume.} The cast solution then undergoes three F/T cycles, each cycle consisting of a 24-hour freeze followed by a 24-hour thaw. This process facilitates physical crosslinking of the PVA matrix through crystallite formation. Once the gel is physically crosslinked, it is transferred to the device.


\subsubsection{Design of Electrode Geometries} 
\label{sec:geometry}

\begin{figure}[t]
	\centering
	\includegraphics[width=0.9\columnwidth]{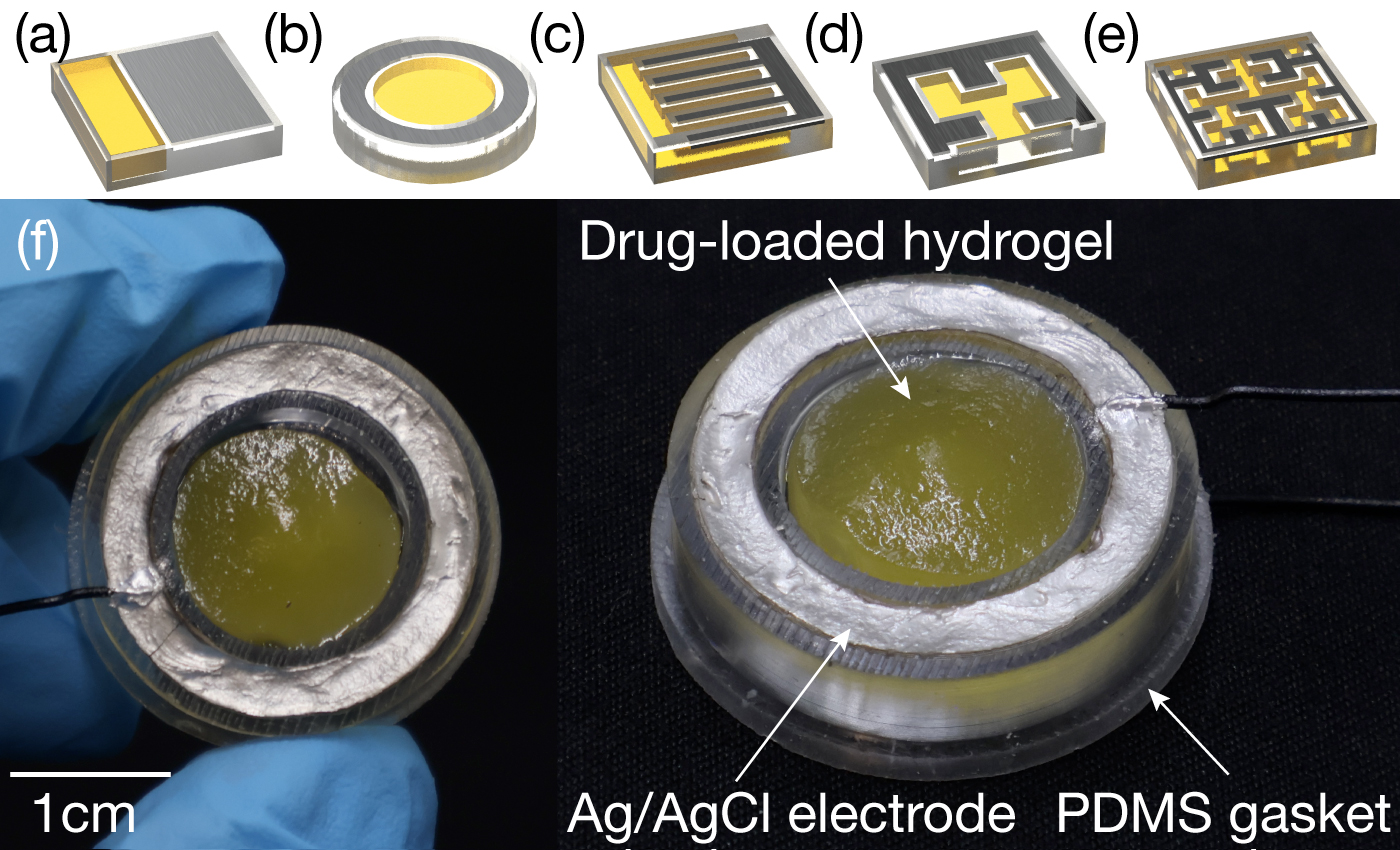}
	\caption{The hydrogel can be fabricated in five geometries: (a) \textit{Plain}, (b) \textit{Circle}, (c) \textit{Interdigital}, (d) \textit{Hilbert 2nd}, and (e) \textit{Hilbert 3rd}. (f) Photo of the device consisting of a drug-loaded hydrogel and an Ag/AgCl electrode housed within a PDMS gasket. \revo{The scale bar represents 1\,cm.}} \label{fig:device}
\end{figure}

To examine how physical crosslinking across different gel geometries influences swelling-mediated microstructural patterns, local electric fields, and, ultimately, drug delivery performance, we designed five gel geometries with a fixed footprint area, but varying curvature and corner density: \textit{Plain}, \textit{Circle}, \textit{Interdigital}, \textit{Hilbert 2nd}, and \textit{Hilbert 3rd} (Fig.~\ref{fig:device}: a, b, c, d, and e, respectively). \revt{\textit{Hilbert 2nd} and \textit{Hilbert 3rd} geometries were directly derived from the fractal Hilbert curve~\cite{Hilbert1891}, while the \textit{Interdigital} geometry choice was motivated by electroadhesive actuator designs that are intended to enhance electric field strength~\cite{guo_optimization_2016}. We also used \textit{Plain} and \textit{Circle} geometries as they are the two most commonly used electrode configurations in commercial iontophoresis systems~\cite{IontoPatch}.} The \textit{Circle} electrode pair was defined by a boundary diameter of 23 mm, while the other geometries were constrained within a 23\,mm $\times$ 23\,mm = 529\,mm$^2$ footprint. The cathodal electrode area (drug-loaded region) was fixed at 175.49\,mm$^2$, and the electrode gap between cathode and anode was maintained at 1 mm across all designs. A parametric geometry-generation pipeline was implemented in CAD software (Grasshopper) to enforce these constraints and ensure consistency across geometries.

\subsubsection{Device Assembly}



Polydimethylsiloxane (PDMS) gaskets were fabricated to house the electrodes and connecting wires (Fig. \ref{fig:device}). The molds for these gaskets were first 3D printed using a resin printer (Standard Resin, Saturn 4 Ultra 16K), post-cured, and subsequently heated at 80 $^\circ$C to remove residual monomers. PDMS (Sylgard 184, Dow Corning) was prepared by mixing the base and curing agent at a 10:0.7 weight ratio (7 wt\% curing agent) and cured at 80 $^\circ$C for 4 hours. After cooling to room temperature, silver/silver chloride (Ag/AgCl) paste was injected into the anode side of the gasket and cured at 100 $^\circ$C for 10 minutes. Following electrode preparation, the prepared drug-loaded hydrogel was thawed and mounted in the cathode compartment of the device. The PDMS gasket served to minimize evaporation and maintain the structural integrity of the hydrogel during the delivery study.

\vspace{-1em}

\subsection{Characterization of Drug-Loaded Gel} 

\subsubsection{Spectral Analysis}

Fourier Transform Infrared Spectroscopy (FTIR) spectra were collected for four sample types: PPy, BSA solution, PPy-PVA hydrogel, and BSA-loaded PPy-PVA hydrogel (Bruker Vertex V80V Vacuum FTIR). Spectra were recorded in absorbance mode over 4000-500 cm$^{-1}$. To improve peak resolution and facilitate identification of overlapping features, second derivative spectra were computed using a Savitzky-Golay filter with a seven-point window. All spectra were processed in MATLAB 2025b, including baseline correction.

\vspace{-1em}

\subsection{Gel Condition Enhancement for Drug Delivery} \label{sec:enhancement}

\begin{figure}[h!] 
	\centering
	\includegraphics[width=\columnwidth]{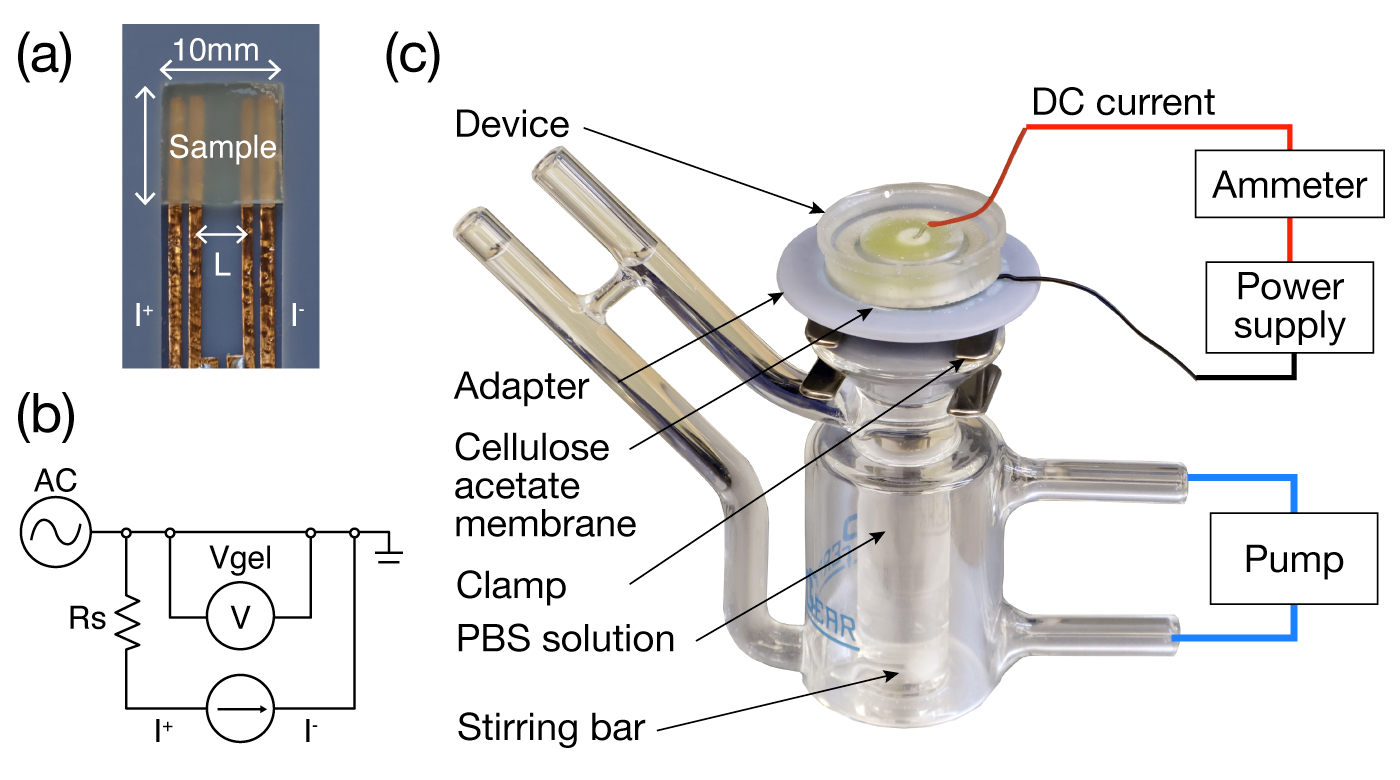}
	\caption{(a) 4-probe measurement circuit. (b) Separate pairs of electrodes are used for current sourcing and voltage sensing. (c) Schematic of the delivery study setup of the BSA-loaded device.} \label{fig:drugdelivery} 
\end{figure}

In this section, we identify the parameters affecting drug delivery performance and employ the one-factor-at-a-time method to identify promising parameters, which are then tested individually. Three samples per condition were prepared and evaluated for all characterization procedures, with the exception of scanning electron microscopy (SEM). For SEM analysis, one sample was imaged for each geometry.


\subsubsection{Conductivity of PPy-PVA Hydrogel} \label{sec:conductivity}

We measured the electrical conductivity of PPy-PVA hydrogels using a four-probe AC method to minimize errors from electrode-gel interface resistance, which is significant in DC measurements (Fig. \ref{fig:drugdelivery}(a)). The conductivity was evaluated as a function of PPy content. Hydrogels were synthesized at nine PPy feed ratios ranging from 10 wt\% to 90 wt\% in 10 wt\% increments. 
For PVA, a 13 wt\% solution was used. Each hydrogel sample had dimensions of 10 mm $\times$ 10 mm $\times$ 2 mm and was subjected to three F/T cycles. A custom test jig with a groove with a matching dimension was fabricated via 3D resin printing (Standard Resin, Saturn 4 Ultra 16K) and subsequently plasma-treated to minimize interference from residual uncured resin (Fig. \ref{fig:drugdelivery}(a)). The test jig was wiped with isopropyl alcohol after every sample. During measurements, a sinusoidal AC signal of 300 Hz and 200 m$V_{rms}$ was applied to the circuitry (Keysight 33600A Series). A series resistor, $R_s$, of 2.2 k\si{\ohm}, was connected in line with the gel sample. The gel voltage, $V_{gel}$, was measured, and the current through the circuit was calculated as, $ I = \frac{V_{rms}-V_{gel}}{R_s}$. The bulk resistance of the gel was then obtained from, $R_{bulk} = \frac{V_{gel}}{I}$ (Fig. \ref{fig:drugdelivery}(b)). Finally, the conductivity, $\sigma$, was calculated using $\sigma = \frac{L}{R_{bulk}A}$, where $A$ is area, 1\,cm$^2$, and $L$ is the linear distance between the inner sensing probe, which was 5 mm (Fig. \ref{fig:drugdelivery}(a)).

\subsubsection{Drug Entrapment Efficiency} \label{sec:ee}

To quantify BSA loading in the PPy-PVA hydrogels, we calculated the entrapment efficiency (EE, \%) as the ratio of the measured mass of encapsulated BSA to the initially loaded mass. Hydrogel samples with 10 mm $\times$ 10 mm $\times$ 2 mm were prepared with 5, 10, 20, 30, 40 concentrations (mg\,mL$^{-1}$) of BSA. The samples were subjected to F/T cycles under the same conditions as used for other characterization experiments. Alongside drug-loaded gels, blank samples with the same dimensions were created and subjected to the same F/T cycles to serve as a baseline. After preparation, each sample was dissolved in 10 mL DI water and vortexed vigorously for 24 hours to ensure complete dispersion. Dissolved samples were analyzed by UV-Vis spectroscopy. 

\subsubsection{Current Density} \label{sec:cd}

To evaluate the impact of current density on the BSA release ($Q$, $\mu$g\,cm$^{-2}$), gel samples measuring 10 mm $\times$ 10 mm $\times$ 2 mm were prepared. 
Current densities of 0.5 mA\,cm$^{-2}$, 1 mA\,cm$^{-2}$, and 2 mA\,cm$^{-2}$ were applied for 30 minutes. During testing, the Ag/AgCl electrode was positioned 1 mm from the drug-loaded gel, and both the electrode and gel were housed in a PDMS gasket and mounted onto a two-compartment Franz diffusion cell (11.28 mm orifice, jacketed, PermeGear). PBS solution from the receptor compartment was sampled and analyzed using UV-Vis spectroscopy. Since the samples contained an unknown contribution of both BSA and PPy, they were processed using the pipeline described in Section \ref{sec:nnls}. \revfi{The current density that resulted in the highest BSA release was used for the drug delivery test.}

\subsection{Coupled Effects of Geometry and Physical Crosslinking on Hydrogel Properties} \label{sec:geom} 

\revo{Hydrogel geometry is defined by a negative mold template, inherently coupling microstructure and electric field distribution. We examine two key properties (microstructure and maximum electric field strength), each affected differently by the F/T driven physical crosslinking process: the microstructure of physically crosslinked hydrogels is characterized via SEM to capture fabrication effects, while electric field strength is evaluated through simulation.}

\subsubsection{Hydrogel Microstructure}


Hydrogel samples with five distinct geometries were fabricated according to the established parameters. The samples underwent identical F/T cycles and were subsequently freeze-dried for 60 hours to preserve their microstructural features. For the three geometries with pronounced periodic geometry features (\textit{Interdigital}, \textit{Hilbert 2nd}, \textit{Hilbert 3rd}; Fig.~\ref{fig:device}(c)-(e)), two distinct local regions were sampled: (i) narrow cross-sections where mechanical constraint is maximal, and (ii) wider sections that experience comparatively lower constraint. In contrast, the \textit{Plain} and \textit{Circle} geometries (Fig.~\ref{fig:device}(a), (b)) consist of a single bulk volume with relatively uniform cross-section. For these samples, we sectioned from the center and near the edge to capture any radial variations that may arise from differential swelling against the mold walls. 
Microstructural features included geometry-level median pore area and bulk porosity, as well as region-specific (local) pore area and porosity.

First, SEM was used to image gel cross-sections (Zeiss Sigma 500). Each image was manually annotated to define a region of interest (ROI) excluding peripheral gel walls 
that lacked discernible pores or exhibited compression artifacts. Pore areas within each ROI were quantified in ImageJ, and local porosity was calculated (MATLAB R2025b) as the pore area fraction. Bulk porosity for each geometry was then computed as the volume-weighted average of regional porosities, using normalized volume fractions extracted from the compiled ImageJ output. 

\subsubsection{Electric Field Distribution}

To evaluate how hydrogel geometry influences the local electric-field distribution, each gel design was modeled in CAD (Grasshopper)\revo{, accounting for consistent volume, and exported as a .STEP file for import into Ansys Maxwell for finite element analysis.} The five electrode geometries were then simulated under a 5 V potential difference to compute the resulting electric field strength. 

\subsection{Drug Delivery Experiments} \label{sec:dd_method}


\subsubsection{Experimental Set Up}
A Franz diffusion cell was modified to accommodate the dimensions of the experimental devices. Custom designed adapters featured either a circular or rectangular diffusion area (Fig.~\ref{fig:drugdelivery}(b)). Each adapter was affixed to the receptor compartment with a clamp, and deviations in receptor volume introduced by the adapter were quantified and incorporated into the concentration calculations. Note that the drug-loaded gel area was kept constant at 175.49\,mm$^{2}$. The donor compartment was replaced with the test device, with the anode and cathode connected to a power supply delivering a constant current density. Current was precisely monitored using a digital ammeter (DC current resolution $\pm$ 0.001\,mA). A cellulose acetate membrane (0.45 $\mu$m, Sterlitech) was clamped between the device and the adapter to serve as the diffusion interface~\cite{wang_CelluloseAcetate_2002, alishahiGlucantimeloadedElectrospunCoreshell2020}. The receptor chamber was filled with 9.85\,mL or 10.26\,mL of phosphate-buffered saline (PBS, pH 7.4) for circular and rectangular adapters, respectively, and maintained at 37 $^\circ$C with continuous stirring. Prior to each experiment, the system was equilibrated for 30 minutes. \revo{Samples from the receptor were collected at 0, 15, 30, 60, 120, and 180 minutes.} 

\subsubsection{Drug Delivery Parameters}

The cumulative amount of BSA that permeated per unit area 
($Q$, $\mu$g\,cm$^{-2}$) was plotted as a function of time to describe permeation kinetics. The pseudo-steady-state region was identified (MATLAB 2025b) by analyzing the time intervals over which the increase in $Q$ was linear by assessing the linearity between successive time intervals, where linearity yielded R$^{2}>$\,0.9570. 
The flux \revf{($J_{ss} = \frac{dQ}{dt}$, $\mu\text{g}\,\text{cm}^{-2}\,\text{h}^{-1}$)} was identified as the slope of the identified region. Lag time ($\tau$, min), which denotes the time required to reach steady state, was estimated by back extrapolating the linear region of the plot to the time axis~\cite{chen_ImprovedApproach_2010, alkilani_TransdermalDrug_2015}. The apparent permeability coefficient ($P_{app}$, cm\,s$^{-1}$), which quantifies membrane permeability to BSA at pseudo-steady-state conditions was calculated as:

\begin{equation}
    \revfi{P_{app} = \frac{J_{ss}}{C_0 \times 3600} = \frac{dQ/dt}{C_0 \times 3600}}
\end{equation}

\noindent where $dQ/dt$ is the flux, $J_{ss}$, at pseudo-steady state, 
and $C_0$ is the initial gel concentration ($\mu$g\,cm$^{-3}$)~\cite{ srinivasan_TransdermalIontophoretic_1989}. The enhancement ratio (ER), which expresses the fold-increase in permeation due to iontophoretic application relative to passive diffusion~\cite{khan_IontophoreticDrug_}, was calculated by dividing the iontophoretic permeability, $P_{app}$ (IP), by passive permeability, $P_{app}$ (P). 
\revt{We performed Spearman rank correlation analysis} \mmrev{using the mean values of the five hydrogel geometries (n\,=\,5; each geometry measured in triplicate)} \revt{to qualitatively assess the relationships between hydrogel properties and the aforementioned drug delivery metrics.}

\section{Results}

\subsection{Characterization of Drug-Loaded Gel} 

\subsubsection{FTIR Data Analysis}

\begin{figure}[h!]
	\centering
	\includegraphics[width=.95\columnwidth]{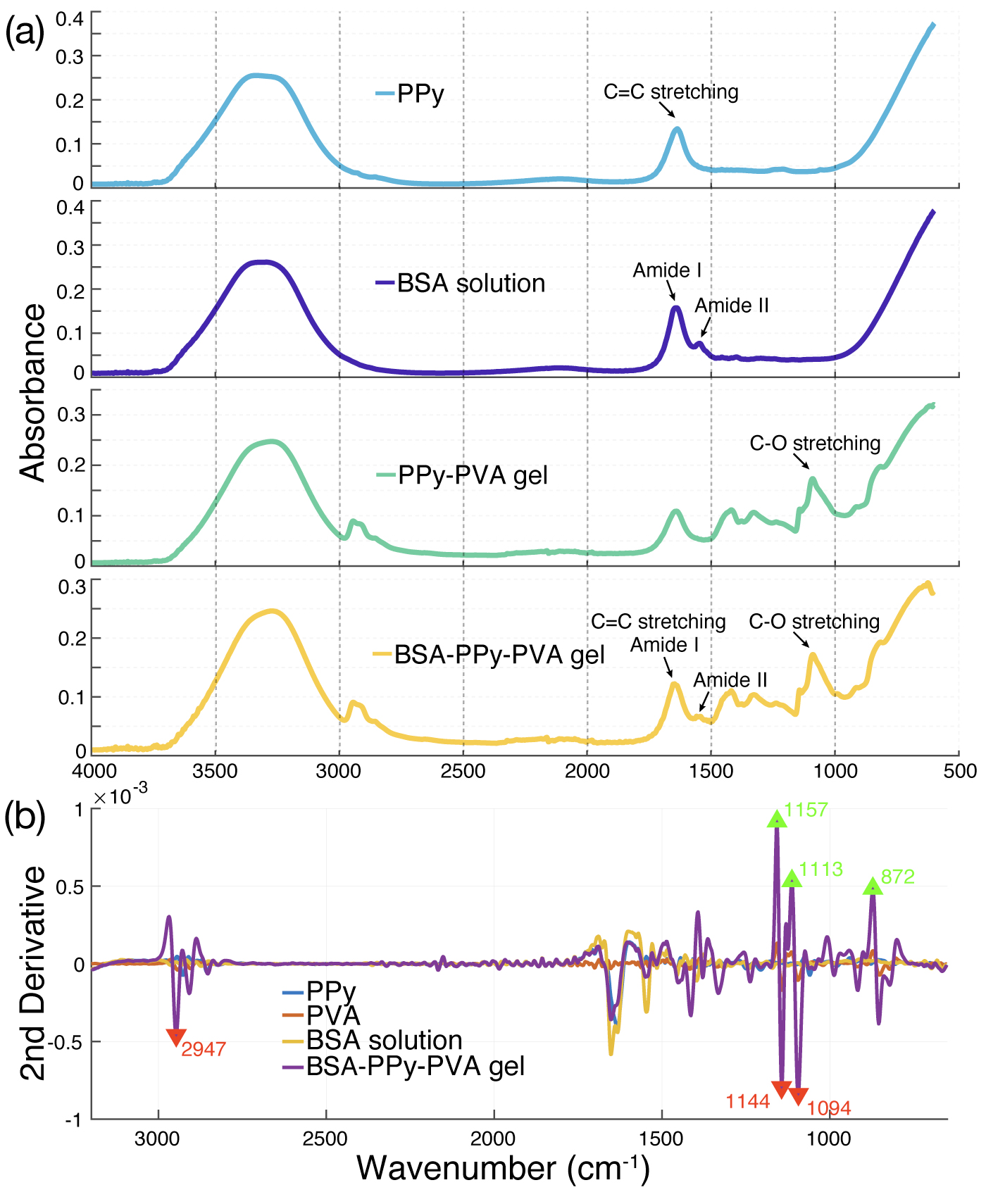}
	\caption{(a) FTIR spectra of PPy, PVA, BSA solution, and the BSA-loaded PPy-PVA gel. (b) Second-derivative FTIR spectra showing the prominent wavenumbers.} 
    \label{fig:ftir}
\end{figure}

The FTIR spectrum of the PPy-PVA hydrogel exhibits characteristic peaks at approximately 2947 cm$^{-1}$, 1089 cm$^{-1}$, and a broad peak centered near 3300 cm$^{-1}$ (Fig.~\ref{fig:ftir}(a)). 2947 cm$^{-1}$ peak corresponds C-H stretching of CH$_{2}$ in the PVA backbone. C-O (and C-O-C) stretching vibrations typical for the PVA backbone~\cite{popescu_2d_nodate, luthfianti_physicochemical_2023, reis_synthesis_2006} at 1085-1090 cm$^{-1}$ range. Finally, the broad absorption band at around 3300 cm$^{-1}$ is assigned to O-H stretching vibrations, originating mainly from hydroxyl groups in PVA and hydrogen-bonded water molecules within the hydrogel~\cite{luthfianti_physicochemical_2023, reis_synthesis_2006}.

\revo{BSA samples exhibit the characteristic amide I peak at 1647 cm$^{-1}$ and amide II at 1546 cm$^{-1}$, which is consistent with previous studies~\cite{kong_fourier_2007, disanto_AdvancementsMidInfrared_2024}, which are observed both in BSA solution (dissolved in PBS) and BSA-PPy-PVA gel samples.} This suggests that BSA retains its native secondary structure upon incorporation into the hydrogel matrix~\cite{mansur_characterization_2004}. 
The second derivative spectra revealed several prominent wavenumbers (Fig.~\ref{fig:ftir}(b)). As the physically crosslinked gel exhibited nuanced peaks across the 700-1500 cm$^{-1}$ region, the applied filter effectively highlighted distinct bands at 2947 cm$^{-1}$, 1144 cm$^{-1}$, 1094 cm$^{-1}$, and 872 cm$^{-1}$. The band at 2947 cm$^{-1}$ corresponds to the asymmetric C-H stretching vibration of the CH$_2$ groups in PVA, confirming the polymer backbone. The peak at 1144 cm$^{-1}$ is associated with C-O stretching in ordered domains, indicative of increased PVA crystallinity following freeze-thaw cycles~\cite{tretinnikov_DeterminationDegree_2012}. The 1094 cm$^{-1}$ band arises from C-O stretching in amorphous PVA segments, while the 872 cm$^{-1}$ peak represents a characteristic PPy vibration attributed to =C-H wagging, corresponding to the ring-stretching mode of the pyrrole structure \cite{xie_SituPolymerization_2019}. The FTIR results indicate that PPy, PVA, and BSA are all incorporated into the gel while retaining their native spectral features.



\vspace{-.3em}
\subsection{Results of Gel Condition Enhancement}

\subsubsection{Conductivity of PPy-PVA Gel} \label{sec:conductivity_result}

\begin{figure}[t]
    \centering
    \includegraphics[width=\linewidth]{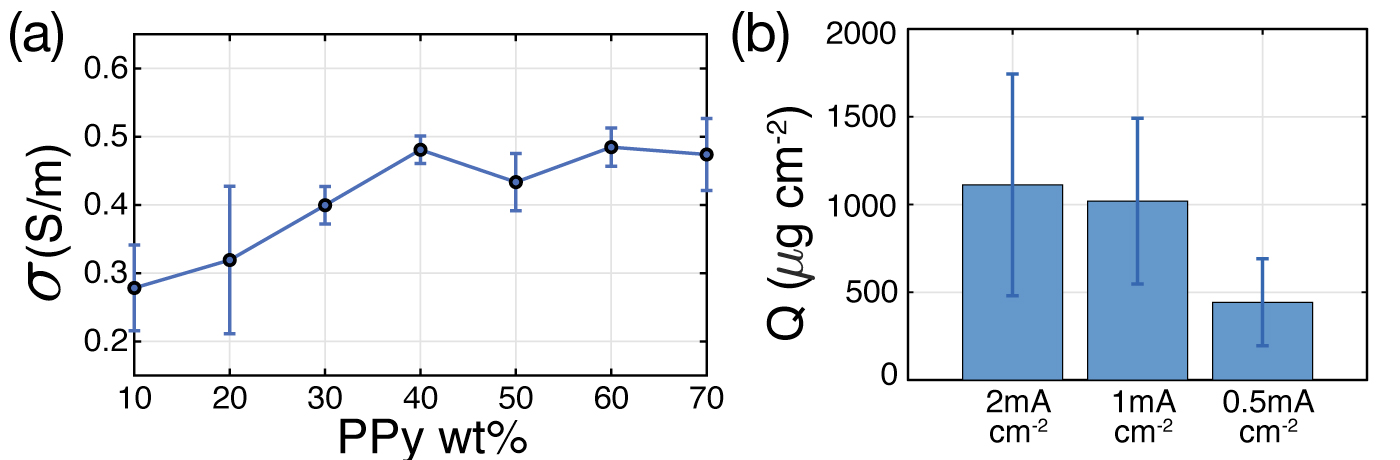}
    \caption{(a) Gel conductivity by PPy concentration. (b) Released BSA across current density groups. (Mean $\pm$ SD).} 
    \label{fig:char}
\end{figure}

80 and 90 wt\% PPy samples did not form stable gels and were therefore excluded from testing. This likely resulted from insufficient viscosity of the precursor mixture, which is necessary for effective physical crosslinking during F/T cycles~\cite{chalmers_increasing_2020}. The 20 wt\% PPy samples exhibited the highest standard deviation in conductivity measurements (Fig.~\ref{fig:char}(a)). This variability may stem from partial phase separation of PPy within the PVA network at this composition, which creates heterogeneous conductive pathways. As a result, even small differences in PPy dispersion or local aggregation can produce measurable fluctuations in conductivity across samples. At lower PPy contents, the continuous PVA matrix dominates and provides uniform gelation~\cite{bercea_recent_2024}, while at higher PPy contents the conductive domains are better interconnected. Based on the testing, we determined that 60 wt\% PPy provides the most favorable balance of high conductivity and relatively low deviation between samples. 

\subsubsection{Drug Entrapment Efficiency} \label{sec:ee_result}

BSA samples prepared with 40 mg\,mL$^{-1}$ did not exhibit gelation and were therefore removed from analysis. This lack of network formation is likely due to the high protein load interfering with PVA chain interactions and crystallite development which disrupted the hydrogen bonding. Among the remaining BSA-loaded gels that successfully formed, the results showed that the entrapment efficiency of BSA peaked at 20 mg\,mL$^{-1}$ as summarized in Table \ref{tab:ee}.

\begin{table}[b]
\centering
\caption{Entrapment efficiency of BSA concentrations. (Mean $\pm$ SD).}
\label{tab:ee}
\small
\renewcommand{\arraystretch}{1.1}
\resizebox{0.7\columnwidth}{!}{
\begin{tabular}{c c}
\hline
\textbf{$C_{\mathrm{BSA}}$ (mg\,mL$^{-1}$)} & 
\textbf{Entrapment Efficiency (\%)} \\ \hline
5  & 11.0 $\pm$ 22.8 \\
10 & 83.7 $\pm$ 57.1 \\
20 & 91.2 $\pm$ 9.0 \\
30 & 35.5 $\pm$ 6.6 \\ \hline
\end{tabular}
}
\end{table}

\subsubsection{Current Density} \label{sec:cd_result}

We applied current densities of 0.5 mA\,cm$^{-2}$, 1 mA\,cm$^{-2}$, and 2 mA\,cm$^{-2}$ samples prepared with 20 mg\,mL$^{-1}$ BSA and 60 wt\% PPy. Since BSA bears a net negative charge at physiological pH, the applied electric field induces electrophoretic migration of BSA molecules towards the anode. As the current density increases, the electrophoretic force on the negatively charged BSA also increases, enhancing its migration (Fig.~\ref{fig:char}(b)). Consequently, samples subjected to 2 mA\,cm$^{-2}$ exhibited the highest quantity of BSA migration, yielding 1112 $\mu$g\,cm$^{-2}$. \revfi{We therefore used 2 mA\,cm$^{-2}$ for the drug delivery test.}

\subsection{Resulting Hydrogel Properties}

\subsubsection{Hydrogel Microstructure}

\begin{figure*}[h]
    \centering
    \includegraphics[width=.93\linewidth]{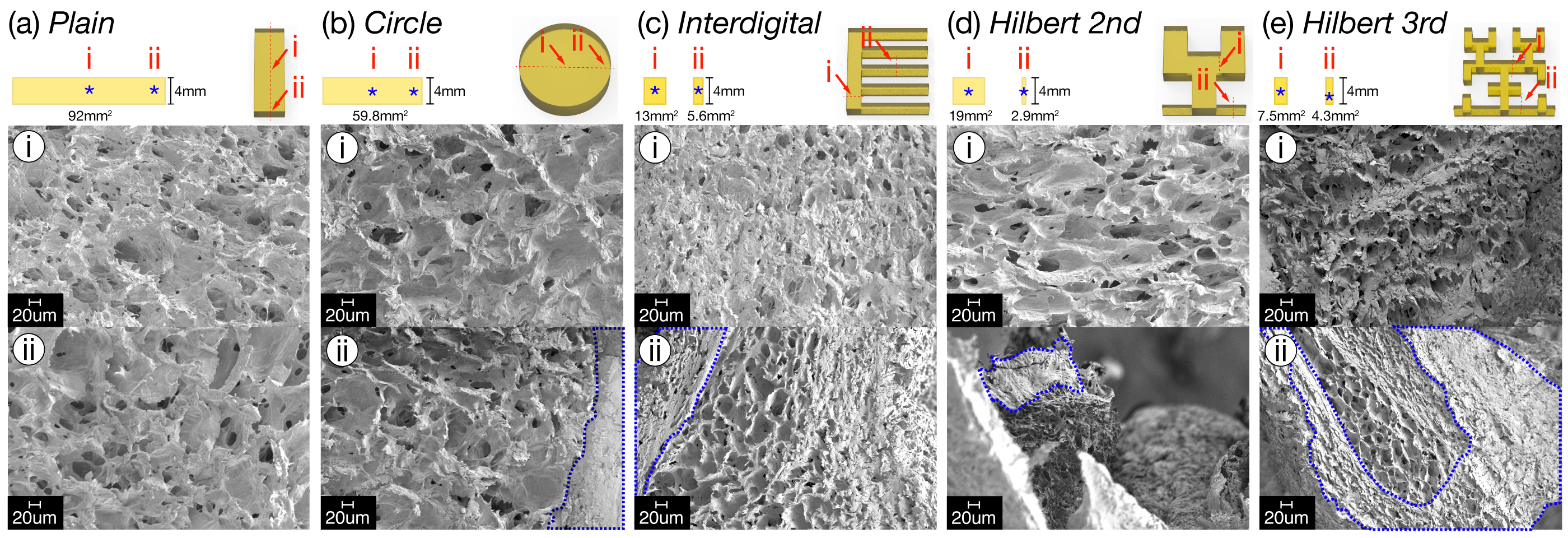}
    \caption{SEM images of BSA-loaded gel samples (500x magnification). Panels (a)-(e) correspond to different geometries, and panels (i)-(ii) represent distinct cross-sectional regions within each geometry. Blue dashed lines denote peripheral gel walls. \revo{Note that the scale bar represents 20 $\mu$m.}}
    \label{fig:SEM}
\end{figure*}

ImageJ analysis revealed substantial variation in median pore area across geometries (Fig.~\ref{fig:SEM}, Table~\ref{tab:porosity}(a)). 
The \textit{Plain} geometry exhibited the largest median pore area (281.66 $\pm$ 357.59~$\mu$m$^2$), followed by \textit{Hilbert 2nd}, \textit{Circle}, \textit{Interdigital}, and \textit{Hilbert 3rd} (234.37 $\pm$ 258.3, 195.06 $\pm$ 241.76, 106.08 $\pm$ 113.97, and 87.75 $\pm$ 98.2~$\mu$m$^2$, respectively). Both \textit{Plain} and \textit{Hilbert 2nd} displayed the greatest within-geometry variability, indicating broad heterogeneity in pore sizes. In contrast, \textit{Hilbert 3rd} showed the smallest dispersion, consistent with a more uniform microstructure.

\begin{table}[t]
\centering
\caption{Summary of microstructure characteristics and electric field strength.}
\scriptsize
\renewcommand{\arraystretch}{1}
\setlength{\tabcolsep}{4.5pt}

\begin{subtable}{\linewidth}
\centering
\caption{Global microstructure and maximum electric field strength by geometry.}
\begin{tabular}{c c c c}
\hline
\textbf{Geometry} & \textbf{Pore Area ($\mu$m$^2$)} & 
\textbf{Bulk Porosity (\%)} & \textbf{Maximum Electric} \\
 &
(Median $\pm$ SD) &
(Mean $\pm$ SD) &
\textbf{Field Intensity} (V/m) \\
\hline
(1) \textit{Plain}  & 281.66 $\pm$ 357.59 & 32.78 $\pm$ 9.68 & 7.5 $\times$ 10$^3$  \\
(2) \textit{Circle} & 195.06 $\pm$ 241.76 & 28.75 $\pm$ 8.25 & 1.14 $\times$ 10$^4$ \\
(3) \textit{Interdigital} & 106.08 $\pm$ 113.97 & 19.23 $\pm$ 3.16  & 1.26 $\times$ 10$^4$ \\
(4) \textit{Hilbert 2nd}  & 234.37 $\pm$ 258.3  & 34.79 $\pm$ 8.78 & 1.32 $\times$ 10$^4$ \\
(5) \textit{Hilbert 3rd}  & 87.75  $\pm$ 98.20   & 30.37 $\pm$ 9.99 & 1.47 $\times$ 10$^4$ \\
\hline
\end{tabular}
\end{subtable}

\vspace{0.6em}

\begin{subtable}{\linewidth}
\centering
\caption{Local microstructure metrics by region.}
\begin{tabular}{c c c c}
\hline
\textbf{Geometry} &
\textbf{Region} &
\textbf{Local Pore Area ($\mu$m$^2$)} &
\textbf{Local Porosity (\%)} \\
&
&
(Median $\pm$ SD) &
(Mean $\pm$ SD) \\

\hline
(1) \textit{Plain}       
& (i)  & 106.57 $\pm$ 177.75 
& 36.00 $\pm$ 0.00 \\
                 
& (ii) & 202.88 $\pm$ 350.81
& 3.75  $\pm$ 0.35 \\
\hline
(2) \textit{Circle}      
& (i)  & 106.57  $\pm$ 177.75
& 31.50 $\pm$ 0.00 \\
                 
& (ii) & 193.05 $\pm$ 179.71
& 4.00 $\pm$ 0.00  \\
\hline
(3) \textit{Interdigital} 
& (i)  & 66.18 $\pm$ 140.81 
& 15.20 $\pm$ 0.00  \\
                 
& (ii) & 41.54 $\pm$ 51.09
& 21.70 $\pm$ 0.00 \\
\hline
(4) \textit{Hilbert 2nd}  
& (i)  & 166.21 $\pm$ 226.43 
& 37.20 $\pm$ 0.00 \\
                 
& (ii) & 12.02 $\pm$ 10.09 
& 2.80 $\pm$ 0.00 \\
\hline
(5) \textit{Hilbert 3rd}  
& (i)  & 92.45 $\pm$ 111.48
&34.40 $\pm$ 2.48\\

& (ii) &18.46 $\pm$ 32.80
& 5.60 $\pm$ 0.32 \\
\hline
\end{tabular}
\end{subtable}

\label{tab:porosity}
\end{table}

Bulk porosity varied among geometries (Table~\ref{tab:porosity}(a)). The \textit{Hilbert 2nd} and \textit{Plain} geometries exhibited the highest bulk porosity (34.79 $\pm$ 8.78\% and 32.78 $\pm$ 9.68\%). The \textit{Interdigital} geometry showed the lowest porosity with minimal variability (19.23 $\pm$ 3.16\%), consistent with tighter packing within the branched channels. Despite consisting of a single continuous domain, the \textit{Circle} geometry exhibited relatively low bulk porosity, while the more intricately partitioned \textit{Hilbert 3rd} geometry showed moderately higher porosity.

Local microstructural analysis further revealed clear region-specific differences in pore area and porosity (Table~\ref{tab:porosity}(b)), consistent with morphological features observed in SEM images (Fig.~\ref{fig:SEM}). In \textit{Plain} and \textit{Circle}, median pore area increased toward the gel edge (region (ii)), suggesting that F/T-induced stress gradients either had limited influence or slightly facilitated pore expansion near the periphery. Supporting this, SEM images of these geometries show relatively large, isotropic pores and more uniform cross-sectional profiles (Fig.~\ref{fig:SEM}(a,b)). In geometries with periodic sub-structures, regional differences were substantially more pronounced. For example, in \textit{Hilbert 2nd}, median pore area decreased sharply from 166.21 $\pm$ 226.43~$\mu$m$^2$ in region (i) to 12.02~$\pm$ 10.09~$\mu$m$^2$ in region (ii), consistent with SEM images (Fig.~\ref{fig:SEM}(d)(i,ii)). Similar reductions in pore area at region (ii) were observed in both \textit{Interdigital} and \textit{Hilbert 3rd}, matching the visible decrease in pore areas in (i) compared with (ii) ((Fig.~\ref{fig:SEM}(c,e)). Local porosity trends paralleled these spatial differences. Geometries with pronounced changes in cross-sectional area, \textit{Plain}, \textit{Circle}, and \textit{Hilbert 2nd}, showed large porosity contrasts between regions (i) and (ii) (36\% vs. 3.75 $\pm$ 0.35\%, 31.5\% vs. 4\%, and 37.2\% vs. 2.8\%, respectively). Geometries with relatively uniform cross-sections, such as \textit{Interdigital}, displayed minimal regional differences in porosity (15.2\% vs. 21.7\%). Unexpectedly, \textit{Hilbert 3rd} showed a pronounced regional disparity (34.4 $\pm$ 2.48\% vs. 5.6 $\pm$ 0.32\%), despite having comparable cross-sectional areas in regions (i) and (ii).

\subsubsection{\revo{Maximum Electric Field Strength}}




 Results from the simulations show that geometries featuring sharp corners or high curvature, such as the \textit{Circle}, \textit{Interdigital}, \textit{Hilbert 2nd} and \textit{Hilbert 3rd} geometries (Table~\ref{tab:porosity}(a)(2,3,4,5)), exhibited localized regions of intensified electric field, commonly referred to as ``hot spots" where electric field lines converge. As a result, the \textit{Hilbert 3rd} geometry generated approximately twice the electric field strength (1.47 $\times$ 10$^4$ V/m) compared to the \textit{Plain} geometry (7.5 $\times$ 10$^3$ V/m). \revfi{Note that the simulation assumes a homogeneous microstructure and applies a fixed 5 V potential difference across the model. This potentiostatic condition was used to isolate the effects of device geometry on the predicted electric potential and electric field distributions. In contrast, the drug delivery experiments were conducted under galvanostatic conditions at a constant current density of 2 mA cm$^{-2}$ to evaluate drug transport kinetics under the experimental delivery condition.}

\vspace{-.3em}


    
    
    
    
    
    
    
    

\subsection{Drug Delivery Through a Cellulose Acetate Membrane} \label{sec:dd_result}

\begin{figure*}[h]
    \centering
    \includegraphics[width=0.95\linewidth]{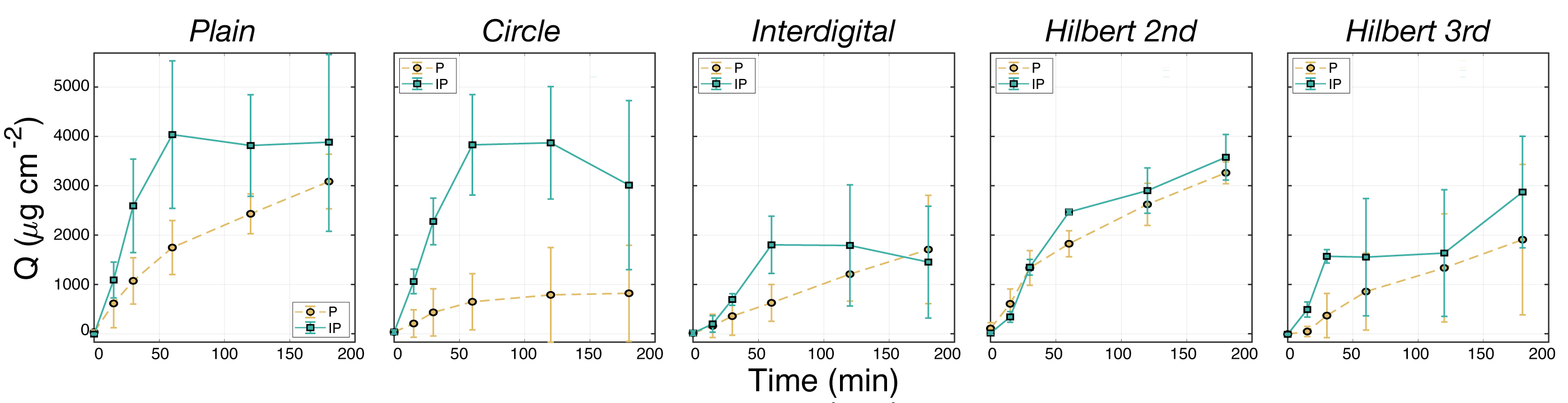}
    \caption{The plots share the same range of $Q$ (y axis). The endpoint of the delivery study was 180 minutes. \revf{Three samples were used for each geometry}. Solid lines represent iontophoretic delivery (IP), and dashed lines represent passive delivery (P).} 
    \label{fig:delivery}
\end{figure*}

Following MATLAB-based linearity analysis, each geometry-delivery condition exhibited a distinct pseudo-linear range consistent with non-uniform permeation kinetics observed in iontophoretic studies (Table~\ref{tab:delivery}, Fig.~\ref{fig:delivery}). The main drug delivery metrics included the cumulative amount of BSA permeated after 3 hours ($Q$), steady-state flux ($J_{ss}$), lag time ($\tau$), apparent permeability coefficient ($P_{app}$), enhancement ratio (ER), and regression coefficient ($R^2$). These metrics are summarized in Table~\ref{tab:delivery}. The delivery condition indicates whether the experiment was conducted under iontophoretic delivery (IP) or passive delivery (P). \revo{We use three samples for each condition of our study, which is consistent with many exploratory in vitro iontophoretic studies that use either artificial membranes or excised skin~\cite{cazares-delgadillo_InvestigationDifferent_2018, kumar_ModulatedIontophoretic_2012, kanikkannanTransdermalIontophoreticDelivery1999, wanasathop_IontophoresisPorcine_2023}. 
The often high standard deviation of the overall delivery can be attributed heterogeneous structures, like that of porcine skin~\cite{cazares-delgadillo_InvestigationDifferent_2018}.}  

\begin{table*}[t!]
    \centering
    \caption{Summary table of drug delivery parameters \mmrev{ \revf{(three samples for each geometry).} For all parameters, uncertainties represent the standard deviation (SD) across n=3 independent replicates. For $J_{ss}$, $\tau$, and $P_{app}$, each replicate value was obtained from independent linear regression of its own steady-state region.}}
    \resizebox{.87\textwidth}{!}{%
    \begin{tabular}{
          >{\centering\arraybackslash}m{1.3cm}  
    c c c c c c c c}
    \hline
    \multirow{2}{*}{\rule{0pt}{1em}\textbf{Delivery}} &
    \multirow{2}{*}{\rule{0pt}{1em}\textbf{Geometry}} &
    \textbf{$Q$ \mmrev{($\mu$g\,cm$^{-2}$)}} &
    \textbf{$J_{ss}$ \mmrev{($\mu\text{g}\,\text{cm}^{-2}\,\text{h}^{-1}$)}} &
    \textbf{$\tau$ \mmrev{(min)}} &
    \textbf{$P_{app}$ \mmrev{(cm\,s$^{-1}$)}} &
    \textbf{ER} &
    \textbf{R$^{2}$} &\\
    & &
    \textbf{\mmrev{(Mean $\pm$ SD)}} &
    \textbf{\revfi{(Mean $\pm$ SD)}} &
    \textbf{\revfi{(Mean $\pm$ SD)}} &
    \textbf{\revfi{(Mean $\pm$ SD)}} &
    &
    \textbf{\mmrev{(Mean $\pm$ SD)}} & \\ \hline
    \vspace{.3em}
    IP & \textit{Plain} &
    $3883 \pm 1.8 \times 10^3$ &
    \revfi{$4602 \pm 1209$} &
    \revfi{$0.209 \pm 0.362$} &
    \revfi{$6.39\times10^{-5} \pm 1.68\times10^{-5}$} &
    \revfi{3.52}
    & $0.9887 \pm 0.0117$ \\

    IP & \textit{Circle} &
    $3012 \pm 1.7 \times 10^3$ &
    \revfi{$3788 \pm 1008$} &
    \revfi{$0.074 \pm 0.128$} &
    \revfi{$5.26\times10^{-5} \pm 1.40\times10^{-5}$} &
    \revfi{6.98}
    & $0.9977 \pm 0.0026$ \\

    IP&\textit{Interdigital}&
    1453 $\pm$ 1.1 $\times 10^3$ &
    \revfi{$1346 \pm 1146$} & 
    \revfi{$5.03 \pm 6.24$} &
    \revfi{$1.87\times10^{-5} \pm 1.59\times10^{-5}$} &
    \revfi{2.25}
    & $0.9890 \pm 0.0135$ \\

    IP &\textit{Hilbert 2nd} &
    3576 $\pm$ 4.6 $\times 10^2$ &
    \revfi{$2052 \pm 920$} &
    \revfi{$1.96 \pm 1.75$} &
    \revfi{$2.85\times10^{-5} \pm 1.28\times10^{-5}$ }&
    \revfi{2.25}
    & $0.9757 \pm 0.0046$ \\

    IP&\textit{Hilbert 3rd}&
    2872 $\pm$ 1.1 $\times 10^3$&
    \revfi{$1891 \pm 434$} &
    \revfi{$0.214 \pm 0.268$} &
    \revfi{$2.63\times10^{-5} \pm 6.03\times10^{-6}$} &
    \revfi{2.90}
    & $0.9929 \pm 0.0123$ \\

    \hline
    \noalign{\vspace{2pt}}
    \revf{Mean~$\pm$~SD}  
    & - & 
    2959 $\pm$ 937 & 
    \revfi{$2736 \pm 1388$} & 
    \revfi{$1.50 \pm 2.12$} &
    \revfi{$(3.80\,\pm \,1.93)\times10^{-5}$} & 
    \revfi{3.58} & 
    $0.9888 \pm 0.0082$\\[.1em]

    \hline\vspace{.5em}
    P & \textit{Plain} &
    3086 $\pm$ 5.5 $\times 10^2$ &
    \revfi{$1306 \pm 817$} &
    $0 \pm 0$ &
    \revfi{$1.81\times10^{-5} \pm 1.14\times10^{-5}$} &
    - & $0.9719 \pm 0.0139$ \\

    P & \textit{Circle} &
    820.8 $\pm$ 9.7 $\times 10^2$&
    \revfi{$543 \pm 342$} &
    \revfi{$3.73 \pm 6.46$} &
    \revfi{$7.54\times10^{-6} \pm 4.76\times10^{-6}$} &
    - & $0.9790 \pm 0.0189$ \\

    P & \textit{Interdigital} &
    1708 $\pm$ 1.1 $\times 10^3$ &
    \revfi{$597 \pm 321$} &
    \revfi{$4.22 \pm 7.31$}  &
    \revfi{$8.30\times10^{-6} \pm 4.45\times10^{-6}$} &
    - & $0.9751 \pm 0.0098$ \\

    P & \textit{Hilbert 2nd} &
    3262 $\pm$ 2.2 $\times 10^2$&
    \revfi{$910 \pm 173$} &
    \revfi{$0 \pm 0$}  &
    \revfi{$1.26\times10^{-5} \pm 2.40\times10^{-6}$} &
    - & $0.9833 \pm 0.0214$ \\

    P & \textit{Hilbert 3rd} &
    1907 $\pm$ 1.5 $\times 10^3$&
    \revfi{$651 \pm 581$} &
    \revfi{$1.64 \pm 2.83$}  &
    \revfi{$9.05\times10^{-6} \pm 8.05\times10^{-6}$} &
    - & $0.977 \pm 0.007$ \\ \hline
    \noalign{\vspace{2pt}}
    \revf{Mean~$\pm$~SD} 
    & - & 2158 $\pm$ 1016 & 
    \revfi{$801 \pm 315$} & 
    \revfi{$1.92 \pm 2.00$} &
    \revfi{$(1.11 \pm 0.44)\times10^{-5}$} &
    - & 
    0.977 $\pm$ 0.004  \\
    [.1em] 
    \hline
    \end{tabular}} \label{tab:delivery}
\end{table*}


\revf{The average cumulative amount, $Q$ ($\mu$g\,cm$^{-2}$), for iontophoretic delivery was $2959 \pm 937$, compared with $2158 \pm 1016$ for passive delivery, suggesting increased permeation under iontophoretic conditions. This observation is consistent with prior studies reporting that iontophoretic stimulation can enhance transdermal drug delivery~\cite{tari_controlled_2021, hasan_IontophoresisBiological_2022}.} 
\revf{The \textit{Hilbert 2nd} geometry exhibited both high simulated electric field intensity and relatively high porosity and was also associated with one of the highest cumulative $Q$ values.} The \textit{Interdigital} geometry was the only geometry where the cumulative delivery in passive delivery exceeded that of iontophoretic delivery.  

During iontophoretic delivery, the \textit{Plain} \revfi{($4602 \pm 1209$)}, \textit{Circle} \revfi{($3788~\pm~1008$)}, and \textit{Hilbert 2nd} \revfi{($2052 \pm 920$)} geometries \revf{had the highest mean $J_{ss}$ ($\mu\text{g}\,\text{cm}^{-2}\,\text{h}^{-1}$) values}, whereas the \textit{Interdigital} and \textit{Hilbert 3rd} geometries exhibited lower \revf{mean} fluxes. \revf{For the \textit{Hilbert 3rd} geometry, the relatively low $J_{ss}$ may be attributable to its shorter linear transport regime, despite its relatively large cumulative amount released. This observation} suggests that much of the drug transport occurred outside the initial linear window, where flux was no longer proportional to time. In practical terms, \camready{\textit{Hilbert 3rd}} may still deliver a substantial total dose over the full experiment, but its release kinetics are less predictable for sustained dosing due to transport rate changes with time. Under passive conditions, \camready{\textit{Plain}} had the highest mean flux (\camready{1306 $\pm$ 817}),  which may be related to its relatively high bulk porosity and comparatively large median pore area (Table~\ref{tab:porosity}(a)).


Across all devices tested, regardless of delivery condition, only minimal lag times, $\tau$ (min), were observed, indicating that drug transport commenced immediately upon initiation of the experiment. The \textit{Interdigital} geometry under iontophoretic conditions exhibited the longest lag time (\revfi{$5.03~\pm~6.24$}~min), which can be attributed to the structural complexity and charge interactions within the gel configuration that likely contribute to this increased delay.


\revf{Additionally, the average $P_{app}$ (cm\,s$^{-1}$) was higher under iontophoretic delivery, \revfi{$(3.80 \, \pm \,$1.93$) \times10^{-5}$}, than under passive delivery, \revfi{$(1.11 \pm 0.44)\times10^{-5}$}. This observation is consistent with the proposed role of an applied electric field in facilitating transdermal transport through mechanisms such as electrorepulsion and electroosmotic flow within the hydrated polymer network.}



\revf{All devices under iontophoresis exhibited ER values greater than 1, which may be attributable to electric field-enhanced membrane permeability and geometry-associated differences in local electric field strength.}



\begin{figure}[t]
    \centering
    \includegraphics[width=\linewidth]{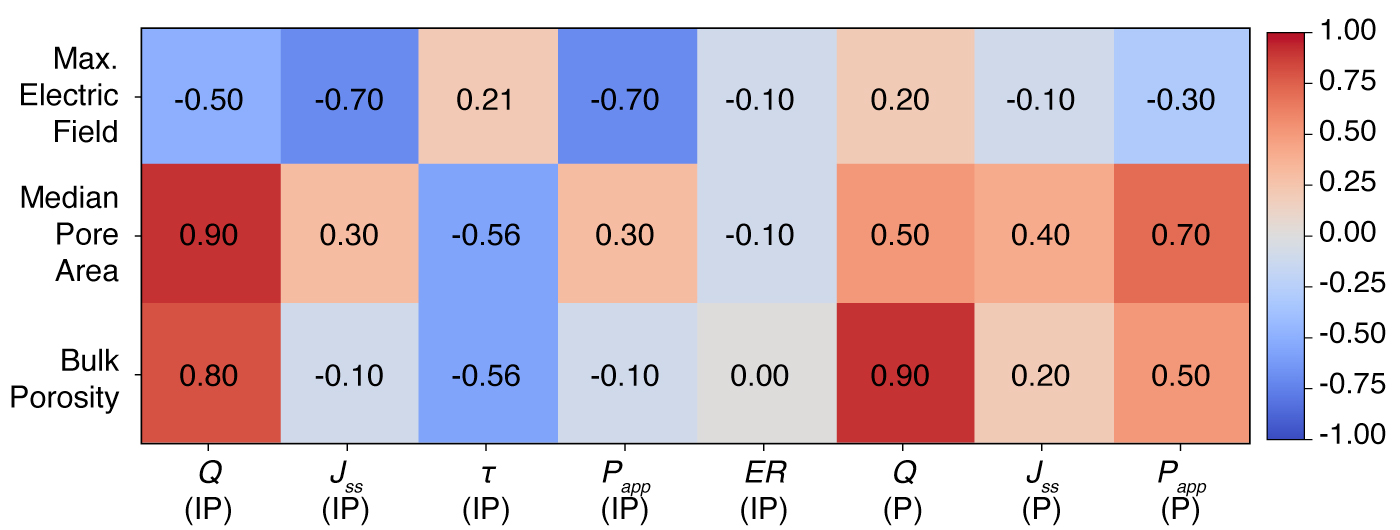}
    \caption{Correlation heatmap of geometry-dependent gel properties to drug delivery outcomes.}
    \label{fig:correlation}
    \vspace{-.3em}
\end{figure}

\mmrev{The reported Spearman coefficients are based on the five geometry-level mean values (Fig.~\ref{fig:correlation}).} 
\mmrev{With n\,=\,5 (denoting the five geometry levels)}, \revo{Spearman correlations are inherently limited and associated with very wide confidence intervals.} 
\revf{Due to the relatively high variability observed, 
as well as the limited number of samples for each geometry, these correlation results should be interpreted as qualitative trends rather than definitive relationships. Thus, although strong associations were observed between some metrics and gel properties, these findings should be considered exploratory.}

\vspace{-.5em}

\section{Discussion}

\subsection{Delivery Performance}

 
\revf{It is worth noting that, on average, the variance across the three samples was relatively high, particularly for passive $Q_P$ and iontophoretic $J_{ss}$, which may reflect variability introduced during the physical crosslinking process. As a result, the observed geometry-associated differences should be interpreted with caution and considered exploratory until confirmed with larger sample sizes.} 
\revt{Also, it is important to note that the findings of this study are based on an artificial membrane model.} \revo{Within this experimental framework, we compare the apparent permeability coefficients, $P_{app}$, under iontophoresis using the hydrogel and artificial membrane.} The average $P_{app}$ under iontophoresis observed in this study was relatively high 
\revfi{$(3.80 \, \pm \,$1.93$) \times10^{-5}$} cm\,s$^{-1}$ \revo{
 compared to values reported in prior iontophoretic studies. For example, such studies have generally reported lower iontophoretic $P_{app}$ values for peptides~\cite{banga_HydrogelBasedLontotherapeutic_1993}. Notably, the $P_{app}$ measured in this work also exceeds values reported for smaller, more mobile molecules such as nicotine~\cite{_IontophoreticTransport_}. To contextualize our results further with prior in vitro studies using excised skin, we consider work that investigated higher-molecular weight drugs delivered from solution. In that study, the permeability coefficients for acetylhexapeptide-3 (molecular weight 888.6 g/mol) and triptorelin acetate (molecular weight 1311 Da) were reported as $4.7\pm \,0.6 \times10^{-6}$ and $3.3 \pm \, 0.8 \times10^{-6}$ cm\,s$^{-1}$, respectively (values converted to match the units of cm s$^{-1}$ used in this work)~\cite{krishnan_IontophoreticSkin_2014}. Both values are lower than the average iontophoretic $P_{app}$ observed in this work, which may be attributed to the use of actual human skin, as opposed to an artificial membrane.} 
\revo{The enhancement ratio (ER) is a useful metric for contextualizing iontophoretic drug delivery studies, however, direct comparisons across reports are limited by differences in current density, skin or membrane type, electrode polarity, and drug molecular weight. For example, an ER of 6.8 was reported for the delivery of acetylhexapeptide-3 (molecular weight 888.6 g/mol) across human skin~\cite{krishnan_IontophoreticSkin_2014}, which is comparable to the ER values observed for the \textit{Hilbert 3rd} and \textit{Plain} geometries in this manuscript, despite differences in formulation and experimental conditions. In contrast, iontophoretic delivery of piroxicam across donated human forearm skin has been reported to yield an ER of 10~\cite{curdy_PiroxicamDelivery_2001}, exceeding the values observed here, likely due to the significantly smaller molecular size of piroxicam.} 
\vspace{-.3em}

\subsection{Limitations of Simulated Electric Field Strength and Scanning Electron Microscopy Results}

\revf{Although the simulation predicts the highest electric field magnitude at the hydrogel corners because of boundary-induced field crowding, these regions are structurally heterogeneous in practice. As shown by the SEM images, dense wall-like surface layers frequently develop near the edges (Fig.~\ref{fig:SEM}), blue-lined regions in (ii)), likely due to dehydration and constrained polymer-chain relaxation during curing. Such interfacial layers may restrict ion mobility and alter local transport pathways and dielectric properties, limiting the effective electric field available for electro-responsive drug transport despite the high simulated field. Therefore, a higher simulated electric field may not necessarily indicate an actually higher effective electric field within the hydrogel.} 
\revo{To reduce heterogeneity from physical crosslinking, chemically crosslinked networks may be used; covalent crosslinking improves stability and structural control, reducing edge effects and the gap between simulated and effective electric fields. Further, future work should quantify effective electric field strength and examine tortuosity and effective permittivity using three-dimensional imaging. For example, X-ray computed tomography (XCT) enables volumetric extraction of tortuosity, porosity, and connectivity to assess bulk heterogeneity and effective permeability, rather than relying on local measurements.}

\revf{One limitation of this study is that the SEM analysis was performed on a single representative specimen for each hydrogel geometry. Although multiple pores were quantified to characterize the pore size distribution within each specimen, these measurements are not independent experimental replicates and therefore do not capture specimen-to-specimen variability. Future studies including multiple independently fabricated specimens per geometry will be needed to evaluate the reproducibility of the observed microstructural characteristics.}

\vspace{-.8em}

\vspace{-.5em}

\subsection{Considerations for In Vivo Study and Real-World Applications}

\revt{In this study, we designed the in vitro current density for maximum drug flux and permeability. However, since this evaluation was done using an artificial membrane, it should not be taken as an in vivo safety threshold. Artificial membranes and excised skin differ substantially in structure and transport properties, so the results should be interpreted as model-specific rather than directly translatable to living skin.} Practical application must account for irritation, epidermal and dermal damage, hydration changes, and safety thresholds, as higher current densities, while enhancing permeation, are associated with increased risk of tissue injury~\cite{prasad_BiophysicalAssessment_2011}.
\revo{The reported values of transport metrics may differ in in vivo studies, primarily because permeation occurs across actual skin with its full biological variability. Nevertheless, ER values are expected to remain a useful comparative measure across studies. For translation of our work to an in vivo animal model, two parameters will require characterization and adjustment: the current density, which should be selected in relation to skin surface impedance, and the electrode spacing, which may influence resistive heating during iontophoresis and therefore affect both safety and transport performance~\cite{ma_effect_2023}.} \revt{In addition to these two parameters, translation of the current hydrogel-based system into practical devices for human use will require further engineering of the adhesive housing to prevent leakage and preserve hydrogel hydration. 
The geometry-dependent electric field distribution is expected to remain significant under practical use conditions, as the PDMS housing preserves the hydrogel geometry and thus the intended field distribution. Accordingly, geometry-induced differences are expected to persist during normal operation, with appreciable changes anticipated only under extreme mechanical deformation.} 

\vspace{-.5em}

\section{Conclusion}

\revo{In this study, we investigate how hydrogel geometry and the physical crosslinking process jointly influence drug transport behavior. \revf{We used a cellulose acetate membrane as a preliminary in vitro model.} To this end, we fabricated a device consisting of drug-loaded PPy-PVA hydrogels as the cathode, an Ag/AgCl anode, and a PDMS gasket. Our PPy-PVA hydrogels were each designed in five distinct geometries to produce unique electric field strengths and microstructures to enable controlled in vitro evaluation of drug transport. For each cathode geometry, we characterized the microstructural properties and simulated electric field strength. Our results show that each hydrogel configuration yields \revfi{different} drug delivery characteristics under both passive diffusion and iontophoretic conditions. \revf{Compared with previous studies using artificial membranes, this work additionally demonstrates geometry-dependent relationships between hydrogel microstructure and drug delivery performance.}} 
These findings provide design guidelines for future gel-based transdermal systems tailored for macromolecular drug transport.

\vspace{-.3em}

\section*{Acknowledgments}

The authors thank Alexandra Lim for her valuable assistance with the BET measurements. The authors also thank the Cornell Center for Materials Research for access to equipment and its staff, John Grazul, Mark A. Pfeifer, and Kevin Silverstein, for their assistance and insights.

\vspace{-.8em}

\bibliographystyle{IEEEtran}
\bibliography{Bibliography}

@article{yang_optimizing_2021,
	title = {Optimizing electrode structure of carbon nanotube gas sensors for sensitivity improvement based on electric field enhancement effect of fractal geometry},
	volume = {11},
	copyright = {2021 The Author(s)},
	issn = {2045-2322},
	doi = {10.1038/s41598-021-96239-1},
	language = {en},
	number = {1},
	urldate = {2025-06-04},
	journal = {Scientific Reports},
	author = {Yang, Taicong},
	year = {2021},
	pages = {16675},
}

@article{guo_optimization_2016,
	title = {Optimization and experimental verification of coplanar interdigital electroadhesives},
	volume = {49},
	copyright = {http://iopscience.iop.org/info/page/text-and-data-mining},
	issn = {0022-3727, 1361-6463},
	doi = {10.1088/0022-3727/49/41/415304},
	language = {en},
	number = {41},
	urldate = {2025-05-22},
	journal = {Journal of Physics D: Applied Physics},
	author = {Guo, J and Bamber, T and Chamberlain, M and Justham, L and Jackson, M},
	year = {2016},
	pages = {415304},
}

@article{tari_controlled_2021,
	title = {Controlled {Transdermal} {Iontophoresis} of {Insulin} from {Water}-{Soluble} {Polypyrrole} {Nanoparticles}: {An} {In} {Vitro} {Study} ({Supp} {Information})},
	volume = {22},
	copyright = {https://creativecommons.org/licenses/by/4.0/},
	issn = {1422-0067},
	shorttitle = {Controlled {Transdermal} {Iontophoresis} of {Insulin} from {Water}-{Soluble} {Polypyrrole} {Nanoparticles}},
	doi = {10.3390/ijms222212479},
	language = {en},
	number = {22},
	urldate = {2025-06-02},
	journal = {International Journal of Molecular Sciences},
	author = {Tari, Kamran and Khamoushian, Soroush and Madrakian, Tayyebeh and Afkhami, Abbas and Łos, Marek Jan and Ghoorchian, Arash and Samarghandi, Mohammad Reza and Ghavami, Saeid},
	year = {2021},
	pages = {12479},
}

@article{kong_fourier_2007,
	title = {Fourier {Transform} {Infrared} {Spectroscopic} {Analysis} of {Protein} {Secondary} {Structures}},
	volume = {39},
	issn = {1672-9145},
	doi = {10.1111/j.1745-7270.2007.00320.x},
	language = {en},
	number = {8},
	urldate = {2025-07-28},
	journal = {Acta Biochimica et Biophysica Sinica},
	author = {Kong, Jilie and Yu, Shaoning},
	year = {2007},
	pages = {549--559},
}

@article{popescu_2d_nodate,
	title = {2D {IR} {Dichroism} {Spectra} of {Poly(Vinyl Alcohol)}},
	language = {en},
	author = {Popescu, Maria-Cristina and Vasile, Cornelia and Simionescu, Bogdan C},
    journal = {Revue Roumaine de Chimie},
    year = {2012},	
    pages = {659-664},
}

@article{luthfianti_physicochemical_2023,
	title = {Physicochemical {Characteristics} and {Antibacterial} {Activities} of {Freeze}-{Thawed} {Polyvinyl} {Alcohol}/{Andrographolide} {Hydrogels}},
	volume = {8},
	issn = {2470-1343},
	doi = {10.1021/acsomega.2c05110},
	number = {3},
	urldate = {2025-07-28},
	journal = {ACS Omega},
	author = {Luthfianti, Halida
Rahmi and Waresindo, William Xaveriano and Edikresnha, Dhewa and Chahyadi, Agus and Suciati, Tri and Noor, Fatimah Arofiati and Khairurrijal, Khairurrijal},
	year = {2023},
	pmid = {36713706},
	pmcid = {PMC9878633},
	pages = {2915--2930},
}

@article{reis_synthesis_2006,
	title = {Synthesis and characterization of poly (vinyl alcohol) hydrogels and hybrids for {rMPB70} protein adsorption},
	volume = {9},
	issn = {1516-1439},
	doi = {10.1590/s1516-14392006000200014},
	language = {en},
	number = {2},
	urldate = {2025-07-28},
	journal = {Materials Research},
	author = {Reis, Elizabeth Fonseca Dos and Campos, Fábia S. and Lage, Andrey Pereira and Leite, Romulo Cerqueira and Heneine, Luiz Guilherme and Vasconcelos, Wander Luiz and Lobato, Zelia Ines Portela and Mansur, Herman Sander},
	year = {2006},
	pages = {185--191},
}

@article{mansur_characterization_2004,
	title = {Characterization of poly(vinyl alcohol)/poly(ethylene glycol) hydrogels and {PVA}-derived hybrids by small-angle {X}-ray scattering and {FTIR} spectroscopy},
	volume = {45},
	copyright = {https://www.elsevier.com/tdm/userlicense/1.0/},
	issn = {0032-3861},
	doi = {10.1016/j.polymer.2004.08.036},
	language = {en},
	number = {21},
	urldate = {2025-07-28},
	journal = {Polymer},
	author = {Mansur, Herman S. and Oréfice, Rodrigo L. and Mansur, Alexandra A.P.},
	year = {2004},
	pages = {7193--7202},
}

@article{arya_quantitative_nodate,
	title = {Quantitative {Estimation} of {Bovine} {Serum} {Albumin} {Protein} {Using} {UV} -{Visible} {Spectroscopy}},
	language = {en},
	author = {Arya, Braham Dutt and Malik, Neetu},
    journal = {World Wide Journal of Multidiscip linary Research and Development},
    year = {2015},
    pages = {66-69},
}

@article{bronze-uhle_synthetic_2016,
	title = {Synthetic nanoparticles of bovine serum albumin with entrapped salicylic acid},
	volume = {Volume 10},
	copyright = {http://creativecommons.org/licenses/by-nc/3.0/},
	issn = {1177-8903},
	doi = {10.2147/NSA.S117018},
	language = {en},
	urldate = {2025-08-04},
	journal = {Nanotechnology, Science and Applications},
	author = {Bronze-Uhle, Erika and Costa, Bruna Carolina and Ximenes, Valdecir Farias and Lisboa-Filho, Paulo Noronha},
	year = {2016},
	pages = {11--21},
}

@article{chalmers_increasing_2020,
	title = {Increasing the {Conductivity} and {Adhesion} of {Polypyrrole} {Hydrogels} with {Electropolymerized} {Polydopamine}},
	volume = {32},
	copyright = {http://pubs.acs.org/page/policy/authorchoice\_termsofuse.html},
	issn = {0897-4756, 1520-5002},
	doi = {10.1021/acs.chemmater.9b03655},
	language = {en},
	number = {1},
	urldate = {2025-08-21},
	journal = {Chemistry of Materials},
	author = {Chalmers, Evelyn and Lee, Haeshin and Zhu, Chuang and Liu, Xuqing},
	year = {2020},
	pages = {234--244},
}

@article{bercea_recent_2024,
	title = {Recent {Advances} in {Poly}(vinyl alcohol)-{Based} {Hydrogels}},
	volume = {16},
	issn = {2073-4360},
	doi = {10.3390/polym16142021},
	number = {14},
	urldate = {2025-08-21},
	journal = {Polymers},
	author = {Bercea, Maria},
	year = {2024},
	pmid = {39065336},
	pmcid = {PMC11281164},
	pages = {2021},
}

@article{ma_effect_2023,
	title = {Effect of geometric parameters of electrodes on skin heating for the design of non‐ablative radiofrequency device},
	volume = {29},
	issn = {0909-752X},
	doi = {10.1111/srt.13472},
	number = {10},
	urldate = {2025-08-25},
	journal = {Skin Research and Technology},
	author = {Ma, Yiyou and Wang, Nianou and Li, Ke and Liang, Huan and Bai, Jingfeng and Ji, Xiang},
	year = {2023},
	pmid = {37881053},
	pmcid = {PMC10560826},
	pages = {e13472},
}

@article{keshava_survey_2003,
	title = {A {Survey} of {Spectral} {Unmixing} {Algorithms}},
	volume = {14},
	language = {en},
	number = {1},
	author = {Keshava, Nirmal},
	year = {2003},
    journal = {Lincoln Library Journal},
}

@article{orth_super-multiplexed_nodate,
	title = {Super-multiplexed fluorescence microscopy via photostability contrast},
	language = {en},
    volume = {9},
    pages = {2943},
    journal = {Biomedical Optics Express},
    year = {2018},
	author = {Orth, Antony and Ghosh, Richik N and Wilson, Emma R and Brown, Hannah and Reineck, Philipp and Gibson, Brant C},
}

@article{pengo_efficient_2013,
	title = {Efficient {Blind} {Spectral} {Unmixing} of {Fluorescently} {Labeled} {Samples} {Using} {Multi}-{Layer} {Non}-{Negative} {Matrix} {Factorization}},
	volume = {8},
	issn = {1932-6203},
	doi = {10.1371/journal.pone.0078504},
	language = {en},
	number = {11},
	urldate = {2025-09-25},
	journal = {PLoS ONE},
	author = {Pengo, Thomas and Muñoz-Barrutia, Arrate and Zudaire, Isabel and Ortiz-de-Solorzano, Carlos},
	editor = {Zuo, Xi-Nian},
	year = {2013},
	pages = {e78504},
}

@article{zobel-roos_process_2017,
	title = {Process {Analytical} {Approach} towards {Quality} {Controlled} {Process} {Automation} for the {Downstream} of {Protein} {Mixtures} by {Inline} {Concentration} {Measurements} {Based} on {Ultraviolet}/{Visible} {Light} ({UV}/{VIS}) {Spectral} {Analysis}},
	volume = {6},
	copyright = {https://creativecommons.org/licenses/by/4.0/},
	issn = {2073-4468},
	doi = {10.3390/antib6040024},
	language = {en},
	number = {4},
	urldate = {2025-09-25},
	journal = {Antibodies},
	author = {Zobel-Roos, Steffen and Mouellef, Mourad and Siemers, Christian and Strube, Jochen},
	year = {2017},
	pages = {24},
}

@article{steiert_photophysical_2018,
	title = {Photophysical {Behavior} of {mNeonGreen}, an {Evolutionarily} {Distant} {Green} {Fluorescent} {Protein}},
	volume = {114},
	issn = {00063495},
	doi = {10.1016/j.bpj.2018.04.013},
	language = {en},
	number = {10},
	urldate = {2025-09-25},
	journal = {Biophysical Journal},
	author = {Steiert, Frederik and Petrov, Eugene P. and Schultz, Peter and Schwille, Petra and Weidemann, Thomas},
	year = {2018},
	pages = {2419--2431},
}

@article{bhatiaEffectModulatedAlternating2014,
	title = {Effect of {Modulated} {Alternating} and {Direct} {Current} {Iontophoresis} on {Transdermal} {Delivery} of {Lidocaine} {Hydrochloride}},
	volume = {2014},
	issn = {2314-6141},
	doi = {10.1155/2014/537941},
	language = {en},
	number = {1},
	urldate = {2025-10-16},
	journal = {BioMed Research International},
	author = {Bhatia, Gaurav and Banga, Ajay K.},
	year = {2014},
	pages = {537941},
}

@article{kastingTheoreticalModelsIontophoretic1992,
	title = {Theoretical models for iontophoretic delivery},
	volume = {9},
	copyright = {https://www.elsevier.com/tdm/userlicense/1.0/},
	issn = {0169409X},
	doi = {10.1016/0169-409X(92)90023-J},
	language = {en},
	number = {2-3},
	urldate = {2025-06-04},
	journal = {Advanced Drug Delivery Reviews},
	author = {Kasting, Gerald B.},
	year = {1992},
	pages = {177--199},
}

@article{costelloIontophoresisApplicationsTransdermal1995,
	title = {Iontophoresis: {Applications} in {Transdermal} {Medication} {Delivery}},
	volume = {75},
	issn = {0031-9023, 1538-6724},
	shorttitle = {Iontophoresis},
	doi = {10.1093/ptj/75.6.554},
	language = {en},
	number = {6},
	urldate = {2025-05-29},
	journal = {Physical Therapy},
	author = {Costello, Charles T and Jeske, Arthur H},
	year = {1995},
	pages = {554--563},
}

@article{zhengIontophoresisdrivenMicroneedlePatch2023,
	title = {Iontophoresis-driven microneedle patch for the active transdermal delivery of vaccine macromolecules},
	volume = {9},
	issn = {2055-7434},
	doi = {10.1038/s41378-023-00515-1},
	language = {en},
	number = {1},
	urldate = {2025-05-29},
	journal = {Microsystems \& Nanoengineering},
	author = {Zheng, Ying and Ye, Rui and Gong, Xia and Yang, Jingbo and Liu, Bin and Xu, Yunsheng and Nie, Gang and Xie, Xi and Jiang, Lelun},
	year = {2023},
	pages = {35},
}

@article{helmyOverviewRecentAdvancements2021,
	title = {Overview of recent advancements in the iontophoretic drug delivery to various tissues and organs},
	volume = {61},
	issn = {17732247},
	doi = {10.1016/j.jddst.2021.102332},
	language = {en},
	urldate = {2025-05-28},
	journal = {Journal of Drug Delivery Science and Technology},
	author = {Helmy, Abdelrahman M.},
	year = {2021},
	pages = {102332},
}

@article{zakzewskiTransdermalDeliveryRegular1998,
	title = {Transdermal delivery of regular insulin to chronic diabetic rats: effect of skin preparation and electrical enhancement},
	volume = {50},
	copyright = {https://www.elsevier.com/tdm/userlicense/1.0/},
	issn = {01683659},
	shorttitle = {Transdermal delivery of regular insulin to chronic diabetic rats},
	doi = {10.1016/S0168-3659(97)00143-0},
	language = {en},
	number = {1-3},
	urldate = {2025-05-13},
	journal = {Journal of Controlled Release},
	author = {Zakzewski, C},
	year = {1998},
	pages = {267--272},
}

@article{pillaiTransdermalDeliveryInsulin2003,
	title = {Transdermal delivery of insulin from poloxamer gel: ex vivo and in vivo skin permeation studies in rat using iontophoresis and chemical enhancers},
	volume = {89},
	copyright = {https://www.elsevier.com/tdm/userlicense/1.0/},
	issn = {01683659},
	shorttitle = {Transdermal delivery of insulin from poloxamer gel},
	doi = {10.1016/S0168-3659(03)00094-4},
	language = {en},
	number = {1},
	urldate = {2025-05-11},
	journal = {Journal of Controlled Release},
	author = {Pillai, O},
	year = {2003},
	pages = {127--140},
}

@article{wangLidocaineLoadedIontophoresisDrivenFiberBased2025,
	title = {Lidocaine-{Loaded} {Iontophoresis}-{Driven} {Fiber}-{Based} {Microneedle} {Patch} for {Controllable} and {Long}-{Lasting} {Transdermal} {Local} {Analgesia}},
	volume = {7},
	issn = {2524-7921, 2524-793X},
	doi = {10.1007/s42765-024-00486-6},
	language = {en},
	number = {1},
	urldate = {2025-05-09},
	journal = {Advanced Fiber Materials},
	author = {Wang, Ting and Shi, Xiaoyu and Wang, Yue and Sun, Hengda and Sun, Yu and Wang, Gang and Jiang, Hong},
	year = {2025},
	pages = {281--295},
}

@article{kanikkannanTransdermalIontophoreticDelivery1999,
	title = {Transdermal iontophoretic delivery of bovine insulin and monomeric human insulin analogue},
	volume = {59},
	copyright = {https://www.elsevier.com/tdm/userlicense/1.0/},
	issn = {01683659},
	doi = {10.1016/S0168-3659(98)00184-9},
	language = {en},
	number = {1},
	urldate = {2025-05-09},
	journal = {Journal of Controlled Release},
	author = {Kanikkannan, N. and Singh, J. and Ramarao, P.},
	year = {1999},
	pages = {99--105},
}

@article{pengIontophoresisIntegratedSmartMicroneedle2024,
	title = {Iontophoresis-{Integrated} {Smart} {Microneedle} {Delivery} {Platform} for {Efficient} {Transdermal} {Delivery} and {On}-{Demand} {Insulin} {Release}},
	volume = {16},
	issn = {1944-8244},
	doi = {10.1021/acsami.4c18381},
	number = {51},
	urldate = {2025-05-09},
	journal = {ACS Applied Materials \& Interfaces},
	author = {Peng, Mingwei and Heng, Ziwen and Ma, Dewei and Hou, Bo and Yang, Keke and Liu, Qinglong and Gu, Zhongwei and Liu, Wei and Chen, Siyuan},
	year = {2024},
	pages = {70378--70391},
}

@article{moraradIontophoresisBasalInsulin2022,
	title = {Iontophoresis of basal insulin controlled delivery based on thermoplastic polyurethane},
	volume = {76},
	issn = {1773-2247},
	doi = {10.1016/j.jddst.2022.103756},
	urldate = {2025-03-21},
	journal = {Journal of Drug Delivery Science and Technology},
	author = {Morarad, Rawita and Naeowong, Witthawat and Niamlang, Sumonman and Sirivat, Anuvat},
	year = {2022},
	pages = {103756},
}

@article{yangSmartphonepoweredIontophoresismicroneedleArray2020,
	title = {Smartphone-powered iontophoresis-microneedle array patch for controlled transdermal delivery},
	volume = {6},
	issn = {2055-7434},
	doi = {10.1038/s41378-020-00224-z},
	language = {en},
	number = {1},
	urldate = {2025-01-07},
	journal = {Microsystems \& Nanoengineering},
	author = {Yang, Jingbo and Li, Yanjun and Ye, Rui and Zheng, Ying and Li, Xiangling and Chen, Yuzhen and Xie, Xi and Jiang, Lelun},
	year = {2020},
	pages = {112},
}

@article{ganguly_BovineSerum_2024,
	title = {Bovine serum albumin as a nanocarrier for efficient encapsulation of hydrophobic garcinol-{A} strategy for modifying the in vitro drug release kinetics},
	volume = {278},
	issn = {01418130},
	doi = {10.1016/j.ijbiomac.2024.134651},
	language = {en},
	urldate = {2025-10-20},
	journal = {International Journal of Biological Macromolecules},
	author = {Ganguly, Shayeri Chatterjee and Mahanti, Beduin and Ganguly, Soumya and Majumdar, Subhabrota},
	year = {2024},
	pages = {134651},
}

@article{aguilera-garrido_ApplicationsSerum_2021,
	title = {Applications of serum albumins in delivery systems: {Differences} in interfacial behaviour and interacting abilities with polysaccharides},
	volume = {290},
	issn = {00018686},
	shorttitle = {Applications of serum albumins in delivery systems},
	doi = {10.1016/j.cis.2021.102365},
	language = {en},
	urldate = {2025-10-20},
	journal = {Advances in Colloid and Interface Science},
	author = {Aguilera-Garrido, Aixa and Del Castillo-Santaella, Teresa and Yang, Yan and Galisteo-González, Francisco and Gálvez-Ruiz, María José and Molina-Bolívar, José Antonio and Holgado-Terriza, Juan Antonio and Cabrerizo-Vílchez, Miguel Ángel and Maldonado-Valderrama, Julia},
	year = {2021},
	pages = {102365},
}

@article{meng_ResearchProgress_2023,
	title = {Research progress on albumin-based hydrogels: {Properties}, preparation methods, types and its application for antitumor-drug delivery and tissue engineering},
	volume = {11},
	issn = {2296-4185},
	shorttitle = {Research progress on albumin-based hydrogels},
	doi = {10.3389/fbioe.2023.1137145},
	language = {en},
	urldate = {2025-10-20},
	journal = {Frontiers in Bioengineering and Biotechnology},
	author = {Meng, Run and Zhu, Huimin and Deng, Peiying and Li, Minghui and Ji, Qingzhi and He, Hao and Jin, Liang and Wang, Bochu},
	year = {2023},
	pages = {1137145},
}

@article{meng_PreparationDrugLoaded_2022,
	title = {Preparation of {Drug}-{Loaded} {Albumin} {Nanoparticles} and {Its} {Application} in {Cancer} {Therapy}},
	volume = {2022},
	issn = {1687-4129},
	doi = {10.1155/2022/3052175},
	language = {en},
	number = {1},
	urldate = {2025-10-20},
	journal = {Journal of Nanomaterials},
	author = {Meng, Run and Zhu, Huimin and Wang, Ziwei and Hao, Shilei and Wang, Bochu},
	year = {2022},
	pages = {3052175},
}

@article{karimi_AlbuminNanostructures_2016,
	title = {Albumin nanostructures as advanced drug delivery systems},
	volume = {13},
	issn = {1742-5247, 1744-7593},
	doi = {10.1080/17425247.2016.1193149},
	language = {en},
	number = {11},
	urldate = {2025-10-20},
	journal = {Expert Opinion on Drug Delivery},
	author = {Karimi, Mahdi and Bahrami, Sajad and Ravari, Soodeh Baghaee and Zangabad, Parham Sahandi and Mirshekari, Hamed and Bozorgomid, Mahnaz and Shahreza, Somayeh and Sori, Masume and Hamblin, Michael R.},
	year = {2016},
	pages = {1609--1623},
}

@article{levin_SwellingConstraints_2023,
	title = {Swelling under {Constraints}: {Exploiting} {3D}‐{Printing} to {Optimize} the {Performance} of {Gel}‐{Based} {Devices}},
	volume = {8},
	issn = {2365-709X, 2365-709X},
	shorttitle = {Swelling under {Constraints}},
	doi = {10.1002/admt.202202136},
	language = {en},
	number = {14},
	urldate = {2025-10-29},
	journal = {Advanced Materials Technologies},
	author = {Levin, Michal and Cohen, Noy},
	year = {2023},
	pages = {2202136},
}

@article{srinivasan_TransdermalIontophoretic_1989,
	title = {Transdermal {Iontophoretic} {Drug} {Delivery}: {Mechanistic} {Analysis} and {Application} to {Polypeptide} {Delivery}},
	volume = {78},
	copyright = {https://www.elsevier.com/tdm/userlicense/1.0/},
	issn = {00223549},
	shorttitle = {Transdermal {Iontophoretic} {Drug} {Delivery}},
	doi = {10.1002/jps.2600780506},
	language = {en},
	number = {5},
	urldate = {2025-11-04},
	journal = {Journal of Pharmaceutical Sciences},
	author = {Srinivasan, V. and Higuchi, W.I. and Sims, S.M. and Ghanem, A.H. and Behl, C.R.},
	year = {1989},
	pages = {370--375},
}

@article{chen_ImprovedApproach_2010,
	title = {An improved approach for determining permeability and diffusivity relevant to controlled release},
	volume = {65},
	issn = {00092509},
	doi = {10.1016/j.ces.2010.08.028},
	language = {en},
	number = {22},
	urldate = {2025-11-04},
	journal = {Chemical Engineering Science},
	author = {Chen, Yiwei and Zhang, Yufeng and Feng, Xianshe},
	year = {2010},
	pages = {5921--5928},
}

@article{alkilani_TransdermalDrug_2015,
	title = {Transdermal {Drug} {Delivery}: {Innovative} {Pharmaceutical} {Developments} {Based} on {Disruption} of the {Barrier} {Properties} of the {Stratum} {Corneum}},
	volume = {7},
	issn = {1999-4923},
	shorttitle = {Transdermal {Drug} {Delivery}},
	doi = {10.3390/pharmaceutics7040438},
	language = {en},
	number = {4},
	urldate = {2025-11-04},
	journal = {Pharmaceutics},
	author = {Alkilani, Ahlam and McCrudden, Maelíosa T. and Donnelly, Ryan},
	year = {2015},
	pages = {438--470},
}

@article{khan_IontophoreticDrug_,
	title = {Iontophoretic drug delivery: {History} and applications},
	language = {en},
	journal = {Journal of Applied Pharmaceutical Science},
    pages = {11-24},
	author = {Khan, Azad and Yasir, Mohd and Asif, Mohd and Chauhan, Iti and Singh, Alok P and Singh, Pradeep and Rai, Shubham},
    year = {2011},
}

@article{disanto_AdvancementsMidInfrared_2024,
	title = {Advancements in {Mid}-{Infrared} spectroscopy of extracellular vesicles},
	volume = {305},
	issn = {13861425},
	doi = {10.1016/j.saa.2023.123346},
	language = {en},
	urldate = {2025-11-07},
	journal = {Spectrochimica Acta Part A: Molecular and Biomolecular Spectroscopy},
	author = {Di Santo, Riccardo and Niccolini, Benedetta and Romanò, Sabrina and Vaccaro, Maria and Di Giacinto, Flavio and De Spirito, Marco and Ciasca, Gabriele},
	year = {2024},
	pages = {123346},
}

@article{xie_SituPolymerization_2019,
	title = {In situ polymerization of polypyrrole on cotton fabrics as flexible electrothermal materials},
	volume = {14},
	issn = {1558-9250, 1558-9250},
	doi = {10.1177/1558925019827447},
	language = {en},
	urldate = {2025-11-07},
	journal = {Journal of Engineered Fibers and Fabrics},
	author = {Xie, Juan and Pan, Wei and Guo, Zheng and Jiao, Shan Shan and Ping Yang, Ling},
	year = {2019},
	pages = {1558925019827447},
}

@article{tretinnikov_DeterminationDegree_2012,
	title = {Determination of the degree of crystallinity of poly(vinyl alcohol) by {FTIR} spectroscopy},
	volume = {79},
	copyright = {http://www.springer.com/tdm},
	issn = {0021-9037, 1573-8647},
	doi = {10.1007/s10812-012-9634-y},
	language = {en},
	number = {4},
	urldate = {2025-11-07},
	journal = {Journal of Applied Spectroscopy},
	author = {Tretinnikov, O. N. and Zagorskaya, S. A.},
	year = {2012},
	pages = {521--526},
}

@article{shah_InnovativeTransdermal_2025,
	title = {Innovative transdermal drug delivery systems: {Benefits}, challenges, and emerging application},
	issn = {2751-7438, 2751-7446},
	shorttitle = {Innovative transdermal drug delivery systems},
	doi = {10.1002/bmm2.70001},
	language = {en},
	urldate = {2025-11-17},
	journal = {BMEMat},
	author = {Shah, Syed Waqas Ali and Li, Xingxing and Yuan, Hao and Shen, Huiling and Quan, Shaohao and Pan, Guifang and Ishfaq, Muhammad and Shah, Abid Ullah and Xie, Hanhan and Shao, Jundong},
	year = {2025},
	pages = {e70001},
}

@article{bouwstra_SkinBarrier_2023,
	title = {The skin barrier: {An} extraordinary interface with an exceptional lipid organization},
	volume = {92},
	issn = {01637827},
	shorttitle = {The skin barrier},
	doi = {10.1016/j.plipres.2023.101252},
	language = {en},
	urldate = {2025-11-17},
	journal = {Progress in Lipid Research},
	author = {Bouwstra, Joke A. and Nădăban, Andreea and Bras, Wim and McCabe, Clare and Bunge, Annette and Gooris, Gerrit S.},
	year = {2023},
	pages = {101252},
}

@incollection{prausnitz_SkinBarrier_,
  author    = {Mark R. Prausnitz and Peter M. Elias and Thomas J. Franz and Matthias Schmuth and Jui-chen Tsai and Gopinathan K Menon and Walter M. Holleran and Kenneth R. Feingold},
  title     = {Skin Barrier and Transdermal Drug Delivery},
  pages = {2065-2073},
  booktitle = {Medical Therapy},
  year      = {2012},
}

@article{elkomy_StructuralOptical_2016,
	title = {Structural and optical properties of pure {PVA}/{PPY} and cobalt chloride doped {PVA}/{PPY} films},
	volume = {9},
	issn = {18785352},
	doi = {10.1016/j.arabjc.2012.04.037},
	language = {en},
	urldate = {2025-11-18},
	journal = {Arabian Journal of Chemistry},
	author = {Elkomy, G.M. and Mousa, S.M. and Abo Mostafa, H.},
	year = {2016},
	pages = {S1786--S1792},
}

@article{pagar_SynthesisPhysical_2023,
	title = {Synthesis, physical, chemical, biological, mechanical and electronic studies of polypyrrole ({PPy}) of versatile scales for electro-mechano, pharmaceutical utilities},
	volume = {9},
	issn = {24058440},
	doi = {10.1016/j.heliyon.2023.e20086},
	language = {en},
	number = {9},
	urldate = {2025-11-18},
	journal = {Heliyon},
	author = {Pagar, Sahebrao B. and Ghorude, Tatyarao N. and Nikolova, Maria P. and SenthilKannan, K.},
	year = {2023},
	pages = {e20086},
}

@article{zhao_NonlinearOptical_2017,
	title = {Nonlinear {Optical} {Properties} of {Novel} {Polypyrrole} {Derivatives} {Bearing} {Different} {Aromatic} {Segments}},
	volume = {08},
	copyright = {http://creativecommons.org/licenses/by/4.0/},
	issn = {2153-117X, 2153-1188},
	doi = {10.4236/msa.2017.811056},
	language = {en},
	number = {11},
	urldate = {2025-11-18},
	journal = {Materials Sciences and Applications},
	author = {Zhao, Wei and Wang, Yun and Wang, Aijian},
	year = {2017},
	pages = {774--783},
}

@article{zhao_HydrogelIonic_2022,
	title = {A {Hydrogel} {Ionic} {Circuit} {Based} {High}‐{Intensity} {Iontophoresis} {Device} for {Intraocular} {Macromolecule} and {Nanoparticle} {Delivery}},
	volume = {34},
	issn = {0935-9648, 1521-4095},
	doi = {10.1002/adma.202107315},
	language = {en},
	number = {5},
	urldate = {2025-11-20},
	journal = {Advanced Materials},
	author = {Zhao, Fan and Fan, Shan and Ghate, Deepta and Romanova, Svetlana and Bronich, Tatiana K. and Zhao, Siwei},
	year = {2022},
	pages = {2107315},
}

@article{wei_ApplicationHydrogels_2023,
	title = {Application of {Hydrogels} in the {Device} of {Ophthalmic} {Iontophoresis}: {Theory}, {Developments} and {Perspectives}},
	volume = {9},
	issn = {2310-2861},
	shorttitle = {Application of {Hydrogels} in the {Device} of {Ophthalmic} {Iontophoresis}},
	doi = {10.3390/gels9070519},
	language = {en},
	number = {7},
	urldate = {2025-11-20},
	journal = {Gels},
	author = {Wei, Dong and Pu, Ning and Li, Si-Yu and Zhao, Na and Song, Zong-Ming and Tao, Ye},
	year = {2023},
	pages = {519},
}

@article{wu_BiomedicalApplications_2024,
	title = {Biomedical applications of stimuli-responsive “smart” interpenetrating polymer network hydrogels},
	volume = {25},
	issn = {25900064},
	doi = {10.1016/j.mtbio.2024.100998},
	language = {en},
	urldate = {2025-11-20},
	journal = {Materials Today Bio},
	author = {Wu, Jiuping and Xue, Wu and Yun, Zhihe and Liu, Qinyi and Sun, Xinzhi},
	year = {2024},
	pages = {100998},
}

@article{han_FlexibleConductive_2022,
	title = {Flexible conductive silk-{PPy} hydrogel toward wearable electronic strain sensors},
	volume = {17},
	issn = {1748-6041, 1748-605X},
	doi = {10.1088/1748-605X/ac5416},
	language = {en},
	number = {2},
	urldate = {2025-11-20},
	journal = {Biomedical Materials},
	author = {Han, Yuanyuan and Sun, Lu and Wen, Chenyu and Wang, Zhaohui and Dai, Jianwu and Shi, Liyang},
	year = {2022},
	pages = {024107},
}

@article{parida_HighlyPorous_2023,
	title = {Highly {Porous} {Polypyrrole} ({PPy}) {Hydrogel} {Support} for the {Design} of a {Co}–{N}–{C} {Electrocatalyst} for {Oxygen} {Reduction} {Reaction}},
	volume = {15},
	copyright = {https://doi.org/10.15223/policy-029},
	doi = {10.1021/acsami.3c08022},
	language = {en},
	number = {31},
	urldate = {2025-11-20},
	journal = {ACS Applied Materials \& Interfaces},
	author = {Parida, Sanjit Kumar and Barik, Tulasi and Chalke, Bhagyashree A. and Amirthapandian, Sankarakumar and Jena, Hrudananda},
	year = {2023},
	pages = {37571--37579},
}

@article{liang_PolyvinylAlcohol_2024,
	title = {Polyvinyl {Alcohol} ({PVA})-{Based} {Hydrogels}: {Recent} {Progress} in {Fabrication}, {Properties}, and {Multifunctional} {Applications}},
	volume = {16},
	issn = {2073-4360},
	shorttitle = {Polyvinyl {Alcohol} ({PVA})-{Based} {Hydrogels}},
	doi = {10.3390/polym16192755},
	language = {en},
	number = {19},
	urldate = {2025-11-20},
	journal = {Polymers},
	author = {Liang, Xiaoxu and Zhong, Hai-Jing and Ding, Hongyao and Yu, Biao and Ma, Xiao and Liu, Xingyu and Chong, Cheong-Meng and He, Jingwei},
	year = {2024},
	pages = {2755},
}

@article{holloway_RoleCrystallization_2013,
	title = {The role of crystallization and phase separation in the formation of physically cross-linked {PVA} hydrogels},
	volume = {9},
	issn = {1744-683X, 1744-6848},
	doi = {10.1039/C2SM26763B},
	language = {en},
	number = {3},
	urldate = {2025-11-20},
	journal = {Soft Matter},
	author = {Holloway, Julianne L. and Lowman, Anthony M. and Palmese, Giuseppe R.},
	year = {2013},
	pages = {826--833},
}

@article{gupta_CombinedEffect_2012,
	title = {A combined effect of freeze--thaw cycles and polymer concentration on the structure and mechanical properties of transparent {PVA} gels},
	volume = {7},
	issn = {1748-605X},
	doi = {10.1088/1748-6041/7/1/015006},
	language = {eng},
	number = {1},
	journal = {Biomedical Materials (Bristol, England)},
	author = {Gupta, Siddhi and Goswami, Sudipta and Sinha, Arvind},

	year = {2012},
	pmid = {22287550},
	pages = {015006},
}

@article{zhang_MixtureAnalysis_2014,
	title = {Mixture analysis using reverse searching and non-negative least squares},
	volume = {137},
	issn = {01697439},
	doi = {10.1016/j.chemolab.2014.06.002},
	language = {en},
	urldate = {2025-11-20},
	journal = {Chemometrics and Intelligent Laboratory Systems},
	author = {Zhang, Zhi-Min and Chen, Xiao-Qing and Lu, Hong-Mei and Liang, Yi-Zeng and Fan, Wei and Xu, Deng and Zhou, Jack and Ye, Fei and Yang, Zheng-Yu},
	year = {2014},
	pages = {10--20},
}

@article{wang_CelluloseAcetate_2002,
	title = {Cellulose acetate membranes for transdermal delivery of scopolamine base},
	volume = {20},
	copyright = {https://www.elsevier.com/tdm/userlicense/1.0/},
	issn = {09284931},
	doi = {10.1016/S0928-4931(02)00018-8},
	language = {en},
	number = {1-2},
	urldate = {2025-11-20},
	journal = {Materials Science and Engineering: C},
	author = {Wang, Fang-Jing and Yang, Yi-Yan and Zhang, Xian-Zheng and Zhu, Xiao and Chung, Tai-shung and Moochhala, Shabbir},
	year = {2002},
	pages = {93--100},
}

@article{alishahiGlucantimeloadedElectrospunCoreshell2020,
	title = {Glucantime-loaded electrospun core-shell nanofibers composed of poly(ethylene oxide)/gelatin-poly(vinyl alcohol)/chitosan as dressing for cutaneous leishmaniasis},
	volume = {163},
	issn = {01418130},
	doi = {10.1016/j.ijbiomac.2020.06.240},
	language = {en},
	urldate = {2025-04-23},
	journal = {International Journal of Biological Macromolecules},
	author = {Alishahi, Mohsen and Khorram, Mohammad and Asgari, Qasem and Davani, Farideh and Goudarzi, Fatemeh and Emami, Amir and Arastehfar, Amir and Zomorodian, Kamiar},
	year = {2020},
	pages = {288--297},
}

@article{prasad_BiophysicalAssessment_2011,
	title = {Biophysical assessment of {DC} iontophoresis and current density on transdermal permeation of methotrexate},
	volume = {1},
	issn = {2230-973X},
	doi = {10.4103/2230-973X.93011},
	language = {en},
	number = {4},
	urldate = {2025-11-20},
	journal = {International Journal of Pharmaceutical Investigation},
	author = {Prasad, Rachna and Koul, Veena and Anand, Sneh},
	year = {2011},
	pages = {234},
}

@article{hasan_IontophoresisBiological_2022,
	title = {Iontophoresis of {Biological} {Macromolecular} {Drugs}},
	volume = {14},
	issn = {1999-4923},
	doi = {10.3390/pharmaceutics14030525},
	language = {en},
	number = {3},
	urldate = {2025-11-20},
	journal = {Pharmaceutics},
	author = {Hasan, Mahadi and Khatun, Anowara and Kogure, Kentaro},
	year = {2022},
	pages = {525},
}

@article{banga_HydrogelBasedLontotherapeutic_1993,
	title = {Hydrogel-{Based} lontotherapeutic {Delivery} {Devices} for {Transdermal} {Delivery} of {Peptide}/{Protein} {Drugs}},
	volume = {10},
	issn = {1573-904X},
	doi = {10.1023/A:1018955631835},
	language = {en},
	number = {5},
	urldate = {2025-11-21},
	journal = {Pharmaceutical Research},
	author = {Banga, Ajay K. and Chien, Yie W.},
	year = {1993},
	pages = {697--702},
}

@article{_IontophoreticTransport_,
	title = {Iontophoretic transport of model compounds from a gel matrix across a cellophane membrane},
	volume = {13},
	number = {14},
	journal = {Drug Development and Industrial Pharmacy},
	publisher = {Taylor \& Francis},
	author = {Bannon, Y. B. and Corish, J. and Corrigan, O. I.},
	year = {1987},
	pages = {2617--2630},
}

@article{curdy_PiroxicamDelivery_2001,
	title = {Piroxicam delivery into human stratum corneum in vivo: iontophoresis versus passive diffusion},
	volume = {76},
	copyright = {https://www.elsevier.com/tdm/userlicense/1.0/},
	issn = {01683659},
	shorttitle = {Piroxicam delivery into human stratum corneum in vivo},
	doi = {10.1016/S0168-3659(01)00418-7},
	language = {en},
	number = {1-2},
	urldate = {2026-04-05},
	journal = {Journal of Controlled Release},
	author = {Curdy, Catherine and Kalia, Yogeshvar N and Naik, Aarti and Guy, Richard H},
	year = {2001},
	pages = {73--79},
}

@article{krishnan_IontophoreticSkin_2014,
	title = {Iontophoretic skin permeation of peptides: an investigation into the influence of molecular properties, iontophoretic conditions and formulation parameters},
	volume = {4},
	copyright = {http://www.springer.com/tdm},
	issn = {2190-393X, 2190-3948},
	shorttitle = {Iontophoretic skin permeation of peptides},
	doi = {10.1007/s13346-013-0181-8},
	language = {en},
	number = {3},
	urldate = {2026-04-05},
	journal = {Drug Delivery and Translational Research},
	author = {Krishnan, Gayathri and Roberts, Michael S. and Grice, Jeffrey and Anissimov, Yuri G. and Moghimi, Hamid R. and Benson, Heather A. E.},
	year = {2014},
	pages = {222--232},
}

@article{cazares-delgadillo_InvestigationDifferent_2018,
	title = {Investigation of {Different} {Iontophoretic} {Currents} {Profiles} for {Short}-{Term} {Applications} in {Cosmetics}},
	volume = {10},
	issn = {1999-4923},
	doi = {10.3390/pharmaceutics10040266},
	language = {en},
	number = {4},
	urldate = {2026-04-05},
	journal = {Pharmaceutics},
	author = {Cázares-Delgadillo, Jennyfer and Planard-Luong, Lien and Gregoire, Sébastian and Serna-Jiménez, César E. and Singhal, Mayank and Kalia, Yogeshvar N. and Merino, Virginia and Merino-Sanjuán, Matilde and Nácher, Amparo and Martínez-Gómez, Ma. Amparo and Burnier-Yalaoui, Veronique and Barbarat, Philippe},
	year = {2018},
	pages = {266},
}

@article{kumar_ModulatedIontophoretic_2012,
	title = {Modulated iontophoretic delivery of small and large molecules through microchannels},
	volume = {434},
	copyright = {https://www.elsevier.com/tdm/userlicense/1.0/},
	issn = {03785173},
	doi = {10.1016/j.ijpharm.2012.05.030},
	language = {en},
	number = {1-2},
	urldate = {2026-04-05},
	journal = {International Journal of Pharmaceutics},
	author = {Kumar, Vijay and Banga, Ajay K.},
	year = {2012},
	pages = {106--114},
}

@article{wanasathop_IontophoresisPorcine_2023,
	title = {Iontophoresis on {Porcine} and {Human} {Gingiva}},
	volume = {40},
	issn = {0724-8741, 1573-904X},
	doi = {10.1007/s11095-023-03535-8},
	language = {en},
	number = {8},
	urldate = {2026-04-08},
	journal = {Pharmaceutical Research},
	author = {Wanasathop, Apipa and Nimmansophon, Patcharawan and Murawsky, Michael and Krishnan, Deepak G. and Li, S. Kevin},
	year = {2023},
	pages = {1977--1987},
}

@article{Hilbert1891,
  author  = {Hilbert, David},
  title   = {Über die stetige Abbildung einer Linie auf ein Flächenstück},
  journal = {Mathematische Annalen},
  year    = {1891},
  volume  = {38},
  pages   = {459--460},
  language = {German}
}

@misc{IontoPatch,
  author = {{IontoPatch}},
  title = {On-the-Go Patch Therapy},
  year = {2026},
  url = {https://iontopatch.com/}
}

\vfill

\clearpage
\twocolumn[{%
\begin{center}
{\Huge\sffamily Geometry-Dependent Properties of \\
Physically Crosslinked PPy-PVA Hydrogels \\ Enable Tunable Iontophoretic Drug Delivery\\ \vspace{1em} \hrule \vspace{1em} Supplementary Material\par}
\vspace{1.0em}
{\lineskip .5em\sffamily\sublargesize Heather Jin Hee Kim,~{\sffamily\textit{Member,~IEEE}}, Mohsen Alishahi, Kyuin Park, Mikaila K. Roncevich, Larissa M. Shepherd, Cara M. Nunez,~{\sffamily\textit{Member,~IEEE}}\par}
\end{center}
\vspace{1.5em}
}]
\setcounter{figure}{0}\renewcommand{\thefigure}{S\arabic{figure}}
\setcounter{table}{0}\renewcommand{\thetable}{S\arabic{table}}
\setcounter{equation}{0}\renewcommand{\theequation}{S\arabic{equation}}
\setcounter{section}{0}\renewcommand{\thesection}{S\arabic{section}}


\section{Contents}

\begin{itemize}
    \item Supplementary Figures S1 to S3
    \item \revfi{Supplementary Table S1}
\end{itemize}



\section{Hilbert-Piano Curves}

\begin{figure}[h]
    \centering
    \includegraphics[width=\linewidth]{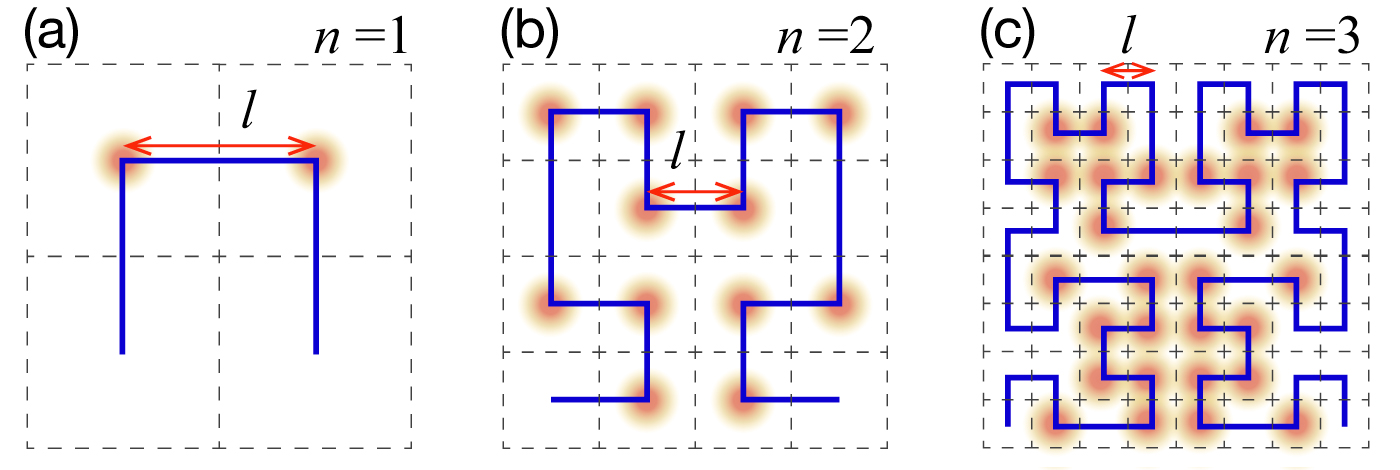}
    \caption{Example of increasing Hilbert curve order: (a) 1st, (b) 2nd, and (c) 3rd order curves. Red dots indicate predicted local “hot spots” where electric field lines concentrate. These hot spots may shift depending on the placement of the other electrode.}
    \label{fig:hilbert}
\end{figure}

In the limit of infinite iterations, a Hilbert curve can fully cover a plane. For a unit area of 1, the length of each segment at the $n^{\text{th}}$ iteration is $l = 1/2^n$, such that the total length at the $n^{\text{th}}$ iteration is $L_n = (4^n - 1)/2^n$. Although the curve remains confined within a unit square, its total length grows exponentially. We illustrate how the Hilbert curve produces increasingly dense space coverage and sharper corner features as it progresses from the first to the third iteration (Fig.~\ref{fig:hilbert}(a-c)).

\newpage


\section{Calibration of BSA and PPy}

\begin{figure}[h!]
    \centering
    \includegraphics[width=\linewidth]{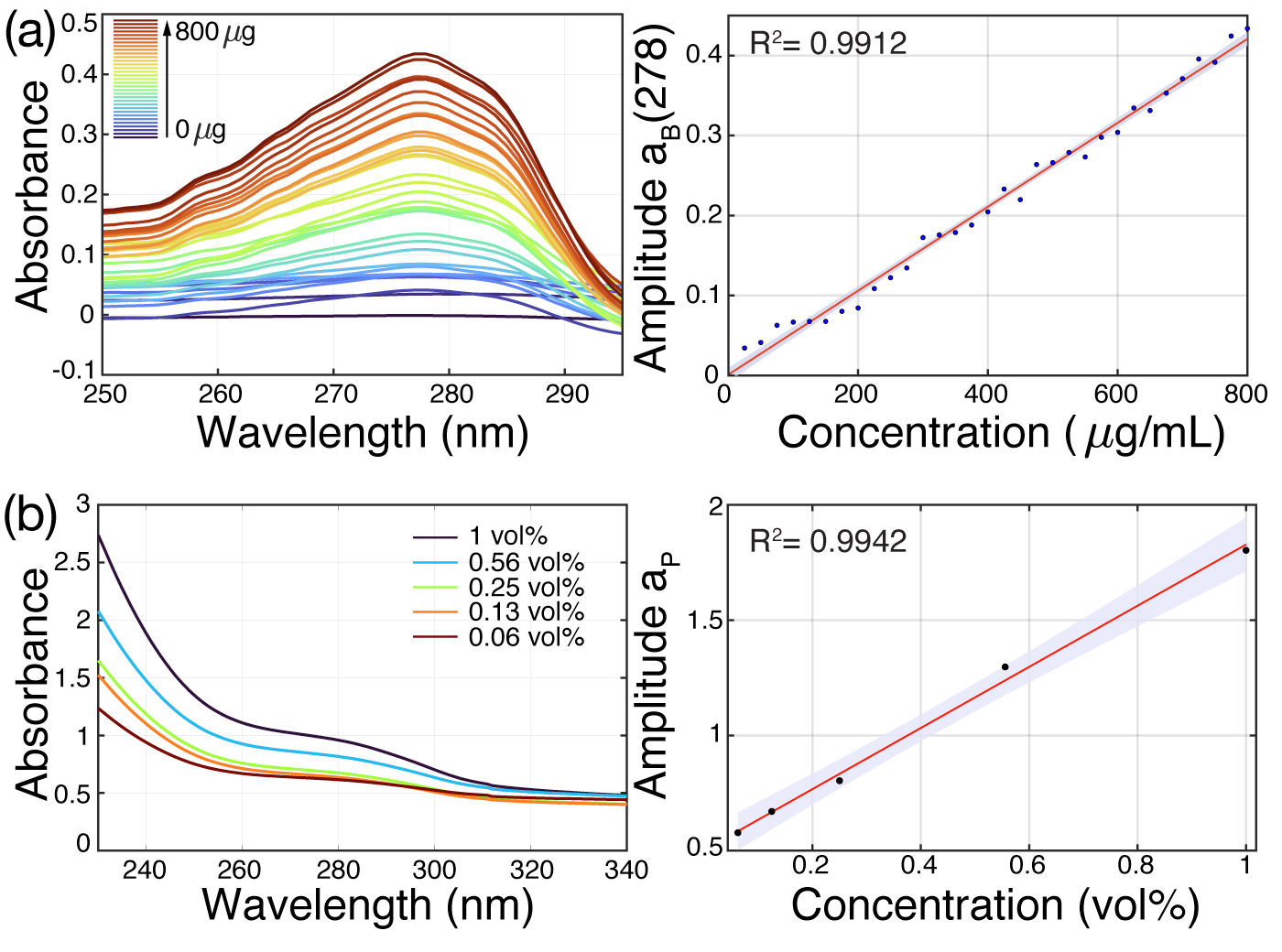}
    \caption{(a) BSA and (b) PPy calibration results. UV-Vis absorbance spectra (left) and NNLS-derived amplitudes $a_B(278), a_P$ plotted as a function of concentration with corresponding linear fits (95\% CI bands are shaded) and $R^2$ values (right).}
    \label{fig:calibration}
\end{figure}

Spectral unmixing of BSA and PPy was performed using NNLS, and the resulting calibration plots are shown in Fig.~\ref{fig:calibration}, with (a) corresponding to BSA and (b) to PPy.

\newpage

\section{Preparation of Drug-Loaded PPy-PVA Hydrogel}

\begin{figure}[h]
    \centering
    \includegraphics[width=\linewidth]{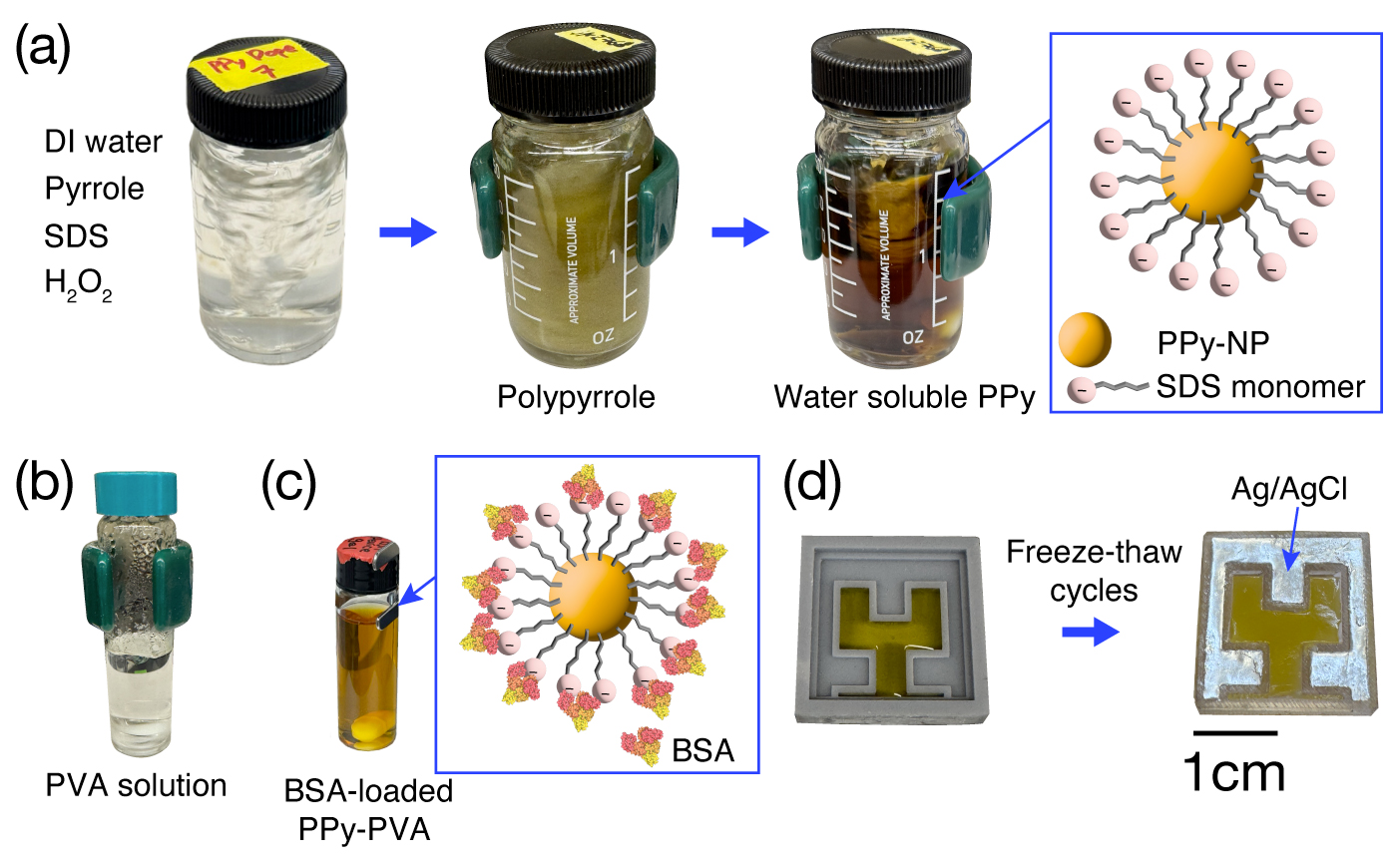}
    \caption{(a) PPy was synthesized by chemical oxidation. (b) A viscous PVA solution was prepared for physical crosslinking and mixed with water-soluble PPy. (c) BSA was added to form the precursor mixture. (d) The solution is freeze-thawed and assembled with an Ag/AgCl electrode.}
    \label{fig:fab}
\end{figure}

Here we provide full images of the preparation steps for the drug-loaded PPy-PVA hydrogel, which followed four stages. First, PPy was synthesized via chemical oxidation (Fig.~\ref{fig:fab}(a)). A viscous PVA solution was then prepared to enable physical crosslinking and blended with water-soluble PPy (Fig.~\ref{fig:fab}(b)). BSA was added to form the precursor polymer mixture (Fig.~\ref{fig:fab}(c)). Finally, the precursor was cast into a resin mold, subjected to freeze-thaw cycles, and assembled with an Ag/AgCl electrode (Fig.~\ref{fig:fab}(d)).

\section{Computing Standard Error (SE)}

\revfi{For each geometry-delivery condition group, BSA concentrations in the receptor were measured at each sampling time and were pooled across samples to calculate the mean and standard deviation at that time point. The mean concentration profile was then converted to a cumulative permeation curve, Q(t). A linear regression was fit to the longest contiguous segment containing at least three time points that satisfied the linearity threshold. $J_{ss}$ was obtained from the fitted slope after conversion from min$^{-1}$ to h$^{-1}$. $P_{app}$ was calculated from $J_{ss}$ using Equation 11. $\tau$ was computed as the negative fitted intercept divided by the slope. Standard error (SE) for $J_{ss}$ and $P_{app}$ were obtained from the regression slope. SE of $\tau$ was obtained by first-order error propagation from the fitted intercept variance, slope variance, and slope-intercept covariance (Table \ref{tab:delivery_SE}).}

\section{Supplementary Code and Data}

The code and data for calculating drug delivery metrics are available in the following online repository: \url{https://github.com/cornell-happi-lab/TBME2026-Transdermal-Drug-Delivery}.

\begin{table*}[t!]
    \centering
    \caption{\revfi{Summary table of drug delivery parameters. (Three samples for each geometry.) For $J_{ss}$, $\tau$, and $P_{app}$, the reported uncertainties represent the standard error of the linear regression used to determine the steady-state region or the propagated standard error of regression-derived quantities.}}
    \resizebox{.87\textwidth}{!}{%
    \begin{tabular}{
          >{\centering\arraybackslash}m{1.3cm}  
    c c c c}
    \hline
    \multirow{2}{*}{\rule{0pt}{1em}\textbf{Delivery}} &
    \multirow{2}{*}{\rule{0pt}{1em}\textbf{Geometry}} &
    \textbf{$J_{ss}$ \mmrev{($\mu\text{g}\,\text{cm}^{-2}\,\text{h}^{-1}$)}} &
    \textbf{$\tau$ \mmrev{(min)}} &
    \textbf{$P_{app}$ \mmrev{(cm\,s$^{-1}$)}} \\
    & &
    \textbf{\mmrev{(Mean $\pm$ SE)}} &
    \textbf{\mmrev{(Mean $\pm$ SE)}} &
    \textbf{\mmrev{(Mean $\pm$ SE)}} \\ \hline
    \vspace{.3em}
    IP & \textit{Plain} &
    $5200 \pm 500$ &
    $0.85 \pm 1.7$ &
    $7.22\times10^{-5} \pm 6.50\times10^{-6}$ \\

    IP & \textit{Circle} &
    $4480 \pm 200$ &
    0 &
    $6.22\times10^{-5} \pm 3.10\times10^{-6}$ \\

    IP & \textit{Interdigital} &
    $2150 \pm 60$ &
    $9.88 \pm 0.86$ &
    $2.98\times10^{-5} \pm 8.10\times10^{-7}$ \\

    IP & \textit{Hilbert 2nd} &
    $555 \pm 70$ &
    0 &
    $7.71\times10^{-6} \pm 9.70\times10^{-7}$ \\

    IP & \textit{Hilbert 3rd} &
    $3160 \pm 700$ &
    $2.00 \pm 3.8$ &
    $4.39\times10^{-5} \pm 9.20\times10^{-6}$ \\

    \hline
    \noalign{\vspace{2pt}}
    \revf{Mean~$\pm$~SD}
    & - & $3109 \pm 1850$ & $2.55 \pm 4.18$ &
    $(4.32 \, \pm \,$\mmrev{2.57})$\times10^{-5}$ \\[.1em]

    \hline\vspace{.5em}
    P & \textit{Plain} &
    $669 \pm 8$ &
    0 &
    $9.29\times10^{-6} \pm 1.20\times10^{-7}$ \\

    P & \textit{Circle} &
    $797 \pm 60$ &
    0 &
    $1.11\times10^{-5} \pm 8.40\times10^{-7}$ \\

    P & \textit{Interdigital} &
    $569 \pm 10$ &
    0 &
    $7.91\times10^{-6} \pm 1.40\times10^{-7}$ \\

    P & \textit{Hilbert 2nd} &
    $850 \pm 40$ &
    0 &
    $1.18\times10^{-5} \pm 6.20\times10^{-7}$ \\

    P & \textit{Hilbert 3rd} &
    $526 \pm 30$ &
    0 &
    $7.30\times10^{-6} \pm 3.80\times10^{-7}$ \\ \hline
    \noalign{\vspace{2pt}}
    \revf{Mean~$\pm$~SD}
    & - & $682 \pm 140$ & $0 \pm 0$ &
    $(9.48 \pm 1.95)\times10^{-6}$ \\[.1em] \hline
    \end{tabular}} \label{tab:delivery_SE}
\end{table*}

\end{document}